\documentclass{article}

\usepackage[main, final]{neurips_2026}

\usepackage[utf8]{inputenc} 
\usepackage[T1]{fontenc}    
\usepackage[hypertexnames=false,hidelinks]{hyperref}       
\usepackage{url}            
\usepackage{booktabs}       
\usepackage{amsfonts}       
\usepackage{nicefrac}       
\usepackage{microtype}      
\usepackage{xcolor}         

\usepackage{natbib} 
\usepackage{color,soul,booktabs}
\setulcolor{blue}
\newcommand{\BibTeX}{\textsc{B\kern-0.1emi\kern-0.017emb}\kern-0.15em\TeX}
 
\usepackage{times}
\usepackage{amsfonts, amssymb, amsmath, amsthm}
\usepackage{bbold}
\usepackage{dsfont}
\usepackage{latexsym}
\usepackage{rotating}
\usepackage{multirow}
\usepackage{xspace}
\usepackage{algorithm,algorithmic}

\usepackage{wrapfig}

\usepackage{tikz}
\usetikzlibrary{positioning,fit,calc}
\usetikzlibrary{arrows.meta}
\usetikzlibrary{calc} 

\usepackage{enumitem}
\setlist[itemize]{leftmargin=1em}

\usepackage{comment}

\newcommand{\bd}[1]{\boldsymbol{#1}}
\newcommand{\set}[1]{\mathcal{#1}}
\newcommand{\1}[1]{\mathbb{1}\big(#1\big)}
\newcommand{\x}{\bd{x}}
\newcommand{\X}{\bd{X}}
\newcommand{\sX}{\bd{\set{X}}}

\newcommand{\Y}{\bd{Y}}
\newcommand{\sY}{\bd{\set{Y}}}
\newcommand{\sS}{\set{S}}
\newcommand{\G}{\set{G}}
\newcommand{\V}{\set{V}}
\newcommand{\W}{\set{W}}
\newcommand{\R}{\set{R}}

\newcommand{\anc}{\mbox{anc}}

\newcommand{\dep}{\mbox{ind}}
\newcommand{\ord}[1]{\bar{#1}}
\usepackage{ stmaryrd }
\newcommand{\event}[1]{\langle #1 \rangle}
\newcommand{\osigma}{\ord{\sigma}}
\newcommand{\opi}{\ord{\pi}}
\newcommand{\otau}{\ord{\tau}}
\newcommand{\orho}{\ord{\rho}}
\newcommand{\oalpha}{\ord{\alpha}}
\newcommand{\obeta}{\ord{\beta}}
\newcommand{\ins}{\mbox{insert}}
\newcommand{\pull}{\mbox{pull}}
\newcommand{\push}{\mbox{push}}
\newcommand{\RIM}{\Delta_{\opi}(\sS^V_n)}
\newcommand{\RSM}{\Delta_{A,B}(\sS^V_n)}
\newcommand{\DeltaG}[1]{\Delta_{#1}(\sX)}
\newcommand{\RRN}{\DeltaG{\G}}

\newcommand{\T}{\bd{\theta}}

\DeclareMathOperator*{\argmax}{arg\,max}
\DeclareMathOperator*{\argmin}{arg\,min}

\usepackage{colortbl}

\definecolor{color123}{RGB}{204,235,197}  
\definecolor{color132}{RGB}{254,217,166}  
\definecolor{color213}{RGB}{179,205,227}  
\definecolor{color231}{RGB}{255,180,193}  
\definecolor{color312}{RGB}{255,250,180}  
\definecolor{color321}{RGB}{222,203,228}  

\newcommand{\co}[2]{\cellcolor{color#1}#2}

\theoremstyle{plain}        
\newtheorem{theorem}{Theorem}[section]
\newtheorem{proposition}[theorem]{Proposition}
\newtheorem{lemma}[theorem]{Lemma}
\newtheorem{corollary}[theorem]{Corollary}

\theoremstyle{definition}   
\newtheorem{definition}[theorem]{Definition}

\theoremstyle{remark}       

\title{Conditional independence and \\graphical models for rankings}

\author{%
  Aritz P\'erez\\
  Basque Center for Applied Mathematics (BCAM)\\
  \texttt{aperez@bcamth.org}}
\hypersetup{pdftitle={Conditional independence and graphical models for rankings},pdfauthor={Aritz P\'erez}}

\begin{document}

\maketitle
\begin{abstract}
Ranking models define probability distributions over rankings and are fundamental in applications such as preference learning and decision-making. Without structural assumptions, these models are intractable due to their factorial complexity. Although independence assumptions are key to reducing complexity and enabling interpretability, standard notions of conditional independence must account for the mutual exclusivity constraints inherent to rankings.
We introduce \emph{relative rank models}, a probabilistic framework based on structured coarsenings of the ranking space that relax these constraints while preserving relative order information. This representation enables a well-defined notion of conditional independence for rankings and allows rankings to be modeled using Bayesian networks. As a result, standard methods for learning and probabilistic inference can be applied to rankings, with interpretable representations based on conditional independence. We further show that two prominent ranking models based on independence assumptions tailored to rankings arise as special cases, providing a unifying framework for independence in ranking models.
\end{abstract}

\section{Introduction}
Modeling rankings is a fundamental problem in machine learning, with applications including preference learning, recommender systems, and information retrieval. In these settings, rankings are represented by permutations and express preferences over a finite set of items~\citep{regenwetter2006socialchoice}, where lower ranks correspond to stronger preferences. A \emph{ranking model} defines a probability distribution over the ranking space and is typically learned from ranking data~\citep{mallows1957ranking}.

The most general ranking model assigns an arbitrary probability to each ranking, requiring a factorial number of parameters in the number of items. This factorial complexity renders such models impractical, as they cannot be learned from limited data, do not admit efficient inference procedures, and lack interpretability.

To overcome these limitations, ranking models commonly rely on structural assumptions that reduce model complexity~\citep{mallows1957ranking,fligner1986distance}. Among these, independence assumptions play a central role. Two representative examples are the \emph{repeated insertion model}~\citep{doignon2004repeated,kenig2018rim} and the \emph{riffle shuffle model}~\citep{huang2009riffled, huang2010learning, huang2012riffled}. The repeated insertion model imposes a fixed insertion order that governs how items are sequentially incorporated into a ranking, whereas the riffle shuffle model introduces a more flexible notion of independence by allowing rankings over disjoint subsets of items to be interleaved.

In probabilistic modeling, the most fundamental notion of independence is \emph{conditional independence}~\citep{dawid1979independence}, which underlies graphical models such as Bayesian networks~\citep{pearl88,lauritzen96}. Graphical models provide compact representations of complex distributions, support efficient inference and learning algorithms, and offer a clear interpretation in terms of conditional independence relations~\citep{koller09}. However, applying conditional independence to rankings requires accounting for the \emph{mutual exclusivity constraints} (MEC) inherent to permutations~\citep{huang2009fourier}, which require distinct items to occupy distinct ranks. These constraints induce strong dependencies among ranking variables and restrict the free parameterization of standard graphical factorizations.


In this work, we address this limitation by shifting the modeling focus from full rankings to structured forms of partial ranking information. Rather than treating partial rankings merely as incomplete observations~\citep{lebanon08kernels, lu2011learning, lan13topk, lu2014effective}, we use them as fundamental modeling primitives~\citep{achab19bucket, chierichetti2018mallows}. By defining structured families of partial rankings, we construct random variables whose realizations induce partitions of the permutation space. This coarsening relaxes the mutual exclusivity constraints and enables a well-defined notion of conditional independence for ranking data, thereby supporting the construction of graphical models for rankings.

\paragraph{Contributions.}
We introduce \emph{relative rank models}, probabilistic models for rankings based on structured coarsenings of the permutation space. Our main contributions are:

\begin{itemize}
    \item We establish fundamental results for \emph{relative rank variables} and the coarsenings of the permutation space, relaxing mutual exclusivity constraints and providing the basis for relative rank models.

    \item We define conditional independence for rankings based on relative rank variables and introduce \emph{relative rank networks}, graphical models for rankings that enable the use of standard theory and algorithms for learning and probabilistic inference, while providing interpretable representations in terms of conditional independence.

    \item We highlight the connection between relative rank networks and existing ranking models, showing that models such as repeated insertion and riffle shuffle can be represented within the RRN framework and that their assumptions correspond to specific sets of conditional independence statements.
\end{itemize}

We provide the proofs of the theoretical results in the main text in Appendix~\ref{app:proofs}.

\paragraph{Notation.}
Let $N=\{1,\dots,n\}$ be a set of indices, and let $V=\{v_1,\dots,v_n\}$ be a set of items. Uppercase letters $A,B,C$ denote subsets of $V$, and $|A|$ denotes the cardinality of $A$. We use $i,j,k \in N$ and $u,v,w \in V$ to denote particular indices and items, respectively. We denote by $\sS^V_n$ the set of bijective functions from $V$ to $N$, called \emph{rankings}, and by $\sS^n_V$ the set of bijective functions from $N$ to $V$, called \emph{orderings}. Rankings are denoted by Greek letters $\rho,\sigma,\tau,\pi$, and their respective orderings by $\orho,\osigma,\otau,\opi$. For simplicity, in examples we identify items with their indices, i.e., $V=N$, and represent rankings and orderings as tuples using $()$ and $[]$, respectively. For example, the ranking $(3,1,2)$ corresponds to the ordering $[2,3,1]$, and vice versa. All directed graphs considered in this work have vertex set $V$ and are identified with their sets of arcs, denoted by uppercase calligraphic letters $\V, \G \subseteq V \times V$.

For a ranking $\sigma\in\sS^V_n$ and a subset $A\subseteq V$, the \emph{partial ranking} $\sigma^A\in\sS^A_{|A|}$ is the bijective function from $A$ to $\{1,\dots,|A|\}$ that preserves the relative order of the items in $A$ under $\sigma$. Its inverse, $\osigma^A\in\sS^{|A|}_A$, is the corresponding \emph{partial ordering}.

\section{Relative rank variables}\label{sec:relative.ranking}
Relative rank variables represent rankings through events in the ranking space. Each realization of such a variable corresponds to a subset of rankings, and together these events form a disjoint cover of the ranking space. Because rankings are permutations, they are subject to mutual exclusivity constraints (MEC), which imply that some combinations of relative ranks cannot occur. In the representation induced by relative rank variables, these constraints appear as empty events. However, carefully designed relative rank variables can break the mutual exclusivity constraints of permutations, so that every realization corresponds to a non-empty event. In this case, the induced events form a partition\footnote{i.e., a disjoint cover of the space using non-empty subsets} of the ranking space. Relative rank variables that break MEC will play a central role in this work, as they allow the definition of ranking models and conditional independence relations without parametric constraints.

Relative rank variables are based on the notion of relative rank.
\begin{definition}[Relative rank]\label{def:relative.rank}
Let $v \in V$ be an item and $P \subseteq V\setminus\{v\}$ be a subset of items called the \emph{reference set}. The \emph{relative rank} of $v$ with respect to $P$ is the mapping
\(
r_P^v : \sS^V_n \to \{1,\dots,|P|+1\},
\)
defined for $\sigma \in \sS^V_n$ as
\[
r_P^v(\sigma) := 1 + \sum_{u \in P} \1{\sigma(u) < \sigma(v)}.
\]
\end{definition}

The value $r_P^v(\sigma)=i$ indicates that item $v$ has the $i$-th smallest rank among the items in $P \cup \{v\}$ under $\sigma$. In particular, $r_P^v(\sigma)=1$ (resp., $r_P^v(\sigma)=|P|+1$) means that $v$ is ranked ahead of (resp., behind) every item in $P$ under $\sigma$. Equivalently, $r_P^v(\sigma)=\sigma^{P \cup \{v\}}(v)$ is the rank of item $v$ in the partial ranking of $P \cup \{v\}$ under $\sigma$, $\sigma^{P \cup \{v\}} \in \sS^{P \cup \{v\}}_{|P|+1}$. For instance, for $\sigma=(2,1,3)$ we have $\sigma^{\{2,3\}}=(1,2)$ and $r^2_{\{3\}}(\sigma)=1$, whereas for $\tau=(2,3,1)$ we have $\tau^{\{2,3\}}=(2,1)$ and $r^2_{\{3\}}(\tau)=2$. 

Another useful interpretation of relative ranks is in terms of insertions into partial orderings. Given a ranking $\sigma$, the value $r^v_P(\sigma)$ specifies the insertion position of item $v$ into the partial ordering $\osigma^{P}$ to obtain $\osigma^{P \cup \{v\}}$. For example, for $\sigma=(2,3,1)$ the relative rank of $1$ with respect to $\{2,3\}$ is $r^{1}_{\{2,3\}}(\sigma)=2$, meaning that inserting item $1$ into $\osigma^{\{2,3\}}=[3,2]$ at position $2$ yields the ordering $\osigma=[3,1,2]$ corresponding to the ranking $\sigma$. Relative ranks generalize pairwise comparisons~\citep{wauthier2013pairwise}, which correspond to reference sets of size $1$.

\begin{definition}[Relative rank variable]\label{def:RRV}
The \emph{relative rank variable} (RRV) for $V$ defined by the reference sets $\{P_v: v \in V\}$ is the $n$-dimensional categorical random variable
\[
\X = (X_1,\dots,X_n),
\qquad
X_v(\sigma) := r^v_{P_v}(\sigma),
\quad \sigma \in \sS^V_n ,
\]
where each component $X_v$ is called the \emph{RRV for $v$}. 
The state space of each variable $X_v$ is
\(
\set{X}_v := \{1,\dots,|P_v|+1\},
\)
and the joint state space of $\X$ is $\sX := \set{X}_1 \times \cdots \times \set{X}_n.$

For a realization $x_v \in \set{X}_v$, the associated event is $\event{x_v} := \{\sigma \in \sS^V_n : X_v(\sigma)=x_v\}$,
and for a joint realization $\x=(x_1,\dots,x_n)\in\sX$, the associated event is
\[
\event{\x}
\;:=\;
\{\sigma \in \sS^V_n : \X(\sigma)=\x\}
\;=\;
\bigcap_{v \in V} \event{x_v}.
\]
We denote by $\event{\sX}=\{\event{\x}: \x \in \sX\}$ the disjoint cover of $\sS^V_n$ induced by $\X$.
\end{definition}

\begin{wrapfigure}{r}{0.50\textwidth}
\vspace{-4.5mm}
\centering

\begin{minipage}[c]{0.45\linewidth}
  \centering
  \includegraphics[height=2.6cm]{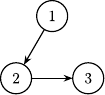}
\end{minipage}
\begin{minipage}[c]{0.45\linewidth}
  \centering
  \setlength{\tabcolsep}{3pt}
  \renewcommand{\arraystretch}{1.0}
  \setlength{\arrayrulewidth}{0.6pt}
  \begin{tabular}{cc|c}
    $\sigma$ & $\osigma$ & $\x$ \\ \hline
    \co{123}{1,2,3} & \co{123}{1,2,3} & \co{123}{1,2,2} \\
    \co{132}{1,3,2} & \co{132}{1,3,2} & \co{132}{1,2,1} \\
    \co{213}{2,1,3} & \co{213}{2,1,3} & \co{213}{1,1,2} \\
    \co{231}{2,3,1} & \co{231}{3,1,2} & \co{132}{1,2,1} \\
    \co{312}{3,1,2} & \co{312}{2,3,1} & \co{213}{1,1,2} \\
    \co{321}{3,2,1} & \co{321}{3,2,1} & \co{321}{1,1,1} \\
  \end{tabular}
\end{minipage}

\caption{%
The variable graph $\V=\{(1,2),(2,3)\}$ defining $\X$ for $V=\{1,2,3\}$. The table shows the mapping of $\X$ from rankings $\sigma \in \sS^V_3$ to the corresponding realizations $\x \in \sX$. Colors in column $\x$ group the rankings belonging to the same event.
}
\label{fig:example.variable.graph}

\vspace{-4mm}
\end{wrapfigure}

An RRV for $V$ maps each ranking to a tuple of relative ranks and induces a disjoint cover of the ranking space through the associated events. Such variables are fully determined by the collection of reference sets $\{P_v : v \in V\}$ defining the relative ranks of each item. This collection of reference sets can be naturally represented by a directed graph, where edges encode which items belong to the reference set of each variable. This graphical representation provides a convenient way to specify RRVs for $V$ and analyze the structural properties of the covers they induce over the ranking space.
\begin{definition}[Variable graph]
Let $\V \subseteq V \times V$ be a directed graph  with $(v,v)\notin\V$ for every $v\in V$. For each $v \in V$, define
\[
P_v^{\V} := \{\, u \in V \setminus \{v\} : (u,v) \in \V \,\}
\]
as the set of parents of node $v$ in $\V$. We say that the RRV $\X=(X_1,\dots,X_n)$ is defined by $\V$ if the reference set of each component $X_v$ is given by $P_v = P_v^{\V}$ for all $v \in V$. The directed graph $\V$ is called the \emph{variable graph} defining $\X$.
\end{definition}

We denote by $\anc(\V)=\{\opi \in \sS^n_V : \pi(u)<\pi(v) \text{ for every } (u,v)\in\V\}$ the set of ancestral (or topological) orderings of $\V$, where $\pi=\opi^{-1}$.

Figure~\ref{fig:example.variable.graph} shows an example of an RRV $\X=(X_1,X_2,X_3)$ for $V$ defined by the variable graph $\V=\{(1,2),(2,3)\}$. The reference sets defined by $\V$ are $P_1=\emptyset$, $P_2=\{1\}$, and $P_3=\{2\}$. The realizations of $\X$ induce the cover of $\sS^V_n$ given by the set of events $\event{\sX}=
\bigl\{
\{(1,2,3)\},
\{(1,3,2),(2,3,1)\},
\{(2,1,3),(3,1,2)\},
\{(3,2,1)\}
\bigr\}.$

Rankings are subject to mutual exclusivity constraints: distinct items must occupy distinct ranks in a ranking. RRVs defined by arbitrary variable graphs can also be subject to MEC, since certain combinations of relative ranks may lead to contradictions and therefore cannot occur simultaneously in any ranking. In the representation induced by an RRV, these constraints appear as empty events in the cover of the ranking space. This motivates identifying RRVs whose realizations correspond only to non-empty events, effectively removing the constraints imposed by permutations. The following definition formalizes this property.

\begin{definition}[Mutual Exclusivity Constraints]\label{def:MEC}
Let $\X$ be an RRV for $V$ with state space $\sX$. We say that $\X$ \emph{is subject to mutual exclusivity constraints} (\emph{MEC}) if there exists a realization $\x \in \sX$ such that $\event{\x} = \varnothing$. If $\event{\x} \neq \varnothing$ for all $\x \in \sX$, we say that $\X$ \emph{breaks MEC}.
\end{definition}

Breaking MEC ensures that every realization of the RRV corresponds to a non-empty subset of rankings. Whether this property holds depends on the structure of the reference sets defining the RRV for $V$, and therefore on the associated variable graph. The following result characterizes when an RRV breaks MEC in terms of the structure of its variable graph.

\begin{theorem}[Breaking MEC]\label{theo:break.MEC}
Let $\X$ be an RRV for $V$ defined by a variable graph $\V$. Then $\X$ breaks MEC if and only if $\V$ is acyclic.
\end{theorem}

When an RRV breaks MEC, every realization corresponds to a non-empty event, and the induced cover $\event{\sX}$ becomes a partition of the ranking space. This property is fundamental for the developments that follow. In particular, RRVs that break MEC allow the definition of probabilistic models and conditional independence relations without structural constraints imposed by permutations.

\begin{wrapfigure}{r}{0.50\textwidth}
\vspace{-4mm}
\centering

\begin{minipage}[c]{0.45\linewidth}
  \centering
  \includegraphics[height=2.6cm]{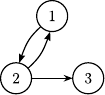}
\end{minipage}
\begin{minipage}[c]{0.45\linewidth}
  \centering
  \setlength{\tabcolsep}{3pt}
  \renewcommand{\arraystretch}{1.0}
  \setlength{\arrayrulewidth}{0.6pt}
  \begin{tabular}{cc|c}
    $\sigma$ & $\osigma$ & $\x$ \\ \hline
    \co{123}{1,2,3} & \co{123}{1,2,3} & \co{123}{1,2,2} \\
    \co{132}{1,3,2} & \co{132}{1,3,2} & \co{132}{1,2,1} \\
    \co{213}{2,1,3} & \co{213}{2,1,3} & \co{213}{2,1,2} \\
    \co{231}{2,3,1} & \co{231}{3,1,2} & \co{132}{1,2,1} \\
    \co{312}{3,1,2} & \co{312}{2,3,1} & \co{213}{2,1,2} \\
    \co{321}{3,2,1} & \co{321}{3,2,1} & \co{321}{2,1,1} \\
  \end{tabular}
\end{minipage}
\vspace{2mm}

\caption{%
The variable graph $\V=\{(1,2),(2,1),(2,3)\}$ contains the cycle $1,2,1$. The realizations $\x \in \{(1,1,1),(1,1,2),(2,2,1),(2,2,2)\}$ also correspond to empty events and do not appear in the table.}
\label{fig:example.cyclic}

\vspace{-2mm}
\end{wrapfigure}

The acyclicity of the variable graph not only guarantees that RRVs break MEC, but also provides a constructive way to generate the rankings in the event associated with each realization. In Appendix~\ref{sec:alg.event}, we present Algorithm~\ref{alg:event} for generating events associated with RRVs defined by acyclic variable graphs. The procedure incrementally constructs partial rankings by inserting elements according to the realization of the RRV, following a topological ordering induced by the variable graph.

Figure~\ref{fig:example.cyclic} shows a cyclic variable graph defining an RRV for $V$ subject to MEC, as predicted by Theorem~\ref{theo:break.MEC}. The presence of the cycle $1,2,1$ induces logical consistency constraints between the RRVs for  items $1$ and $2$. As a consequence, the realizations $\x_{\{1,2\}} \in \{(1,1),(2,2)\}$ have associated empty events. Thus the joint realizations $\x \in \{(1,1,1), (1,1,2), (2,2,1), (2,2,2)\}$ are contradictory and therefore correspond to empty events. For additional examples of variable graphs, see Appendix~\ref{app:variable.graphs}.

\section{Relative rank models}\label{sec:RRM}
Once an RRV for $V$ is defined, probabilistic models over rankings can be constructed by assigning probabilities to its realizations. Such models are called relative rank models. In this section, we formally introduce relative rank models and describe their relationship with ranking models through the pushforward mapping.

\begin{definition}[Relative rank model]
Let $\X$ be an RRV for $V$. A \emph{relative rank model} (\emph{RRM}) is a probability distribution
\[
h \in \Delta(\sX).
\]
\end{definition}

An RRM assigns probability mass to the realizations of the RRV. Under conditions concerning empty events specified below, this assignment defines a probability distribution over the non-empty events of the disjoint cover $\event{\sX}$ induced by $\X$.

Next, we connect RRMs with ranking models through the pushforward relation.

\begin{definition}[Pushforward RRM]\label{def:push.pull}
Let $\X$ be an RRV for $V$. The pushforward mapping under $\X$ is the function $\push_{\X}:\Delta(\sS^V_n)\to \Delta(\sX)$, such that $h=\push_{\X}(p)$ for any ranking model $p$, where
\begin{equation}
h(\x) := p(\x)= \sum_{\sigma \in \event{\x}} p(\sigma),
\qquad \x \in \sX.\label{eq:push}
\end{equation}
In addition, the \emph{pullback} of $h$ under $\X$ is the set of ranking models satisfying Eq.~\ref{eq:push}, denoted by $\pull(h) \subseteq \Delta(\sS^V_n)$.  
\end{definition}

The pushforward relation can be used to characterize the set of RRMs associated with ranking models.

\begin{definition}[Admissible RRM]\label{def:admissible}
Let $\X$ be an RRV for $V$. We say that an RRM $h$ for $\X$ is \emph{admissible} if there exists $p \in \Delta(\sS^V_n)$ such that $\push_{\X}(p)=h$. Otherwise, it is \emph{inadmissible}.
\end{definition}


Admissible RRMs are those that correspond to at least one ranking model through the pullback relation. By construction, any RRM obtained as the pushforward of a ranking model is admissible. However, RRMs can also be defined directly over $\sX$, and in this case admissibility is not guaranteed.

Admissibility imposes a structural constraint: an RRM must assign zero probability to all realizations associated with empty events. This condition is automatically satisfied when the underlying RRV breaks MEC, since all realizations correspond to non-empty events. In contrast, when the RRV is subject to MEC, admissibility requires enforcing structural zeros. A detailed characterization is provided in Appendix~\ref{sec:discussion.admissible}.


\section{Conditional independence for rankings}

Although RRMs reduce the complexity of ranking models by controlling the size of the reference sets $P_v$ for $v \in V$, the number of free parameters of a general RRM is given by
\begin{equation}
\left(\prod_{v \in V} (|P_v|+1)\right)-1.\label{eq:params.free}
\end{equation}

As in classical probabilistic modeling, a natural strategy to further reduce model complexity is to impose conditional independence assumptions and factorize the model accordingly. The challenge in the ranking setting is not merely to impose independence, but to define it in a representation that is structurally compatible with rankings. RRVs provide such a representation, allowing conditional independence relations to be defined analogously to multidimensional categorical variables.

\begin{definition}[Conditional independence]\label{def:conditional.independence}
Let $A, B \subseteq V$ be disjoint non-empty subsets, let $C \subseteq V \setminus (A \cup B)$, and let $\X$ be an RRV for $V$ with an RRM $h$. We say that $\X_A$ is conditionally independent of $\X_B$ given $\X_C$ under $h$, denoted by $\X_A \perp_h \X_B \mid \X_C$, if
\begin{equation}
h(\x_{A,B}\mid \x_{C})= h(\x_{A}\mid \x_{C})h(\x_{B}\mid \x_{C}),\label{eq:cond.indep}
\end{equation}
for all $(\x_A, \x_B, \x_C) \in \sX_A \times \sX_B \times \sX_C$ whenever $h(\x_{C})>0$. Let $p\in \Delta(\sS^V_n)$ be a ranking model. We say that $\X_A$ is conditionally independent of $\X_B$ given $\X_C$ under $p$, denoted by $\X_A \perp_p \X_B \mid \X_C$, if the pushforward of $p$ under $\X$ satisfies Eq.~\ref{eq:cond.indep}. When $C = \varnothing$, these relations reduce to (unconditional) independence, denoted by $\X_A \perp_h \X_B$ and $\X_A \perp_p \X_B$, respectively.
\end{definition}

In words, conditional independence indicates that knowing a realization of $\X_B$ does not provide any additional information about the probabilities of $\X_A$ once the realization of $\X_C$ is known.

In general, given an ordering of the items $\opi$, any RRM can be factorized using the chain rule as
$h(\x)=\prod_{v \in V} h(x_v \mid \x_{<_{\pi} v})$
where $<_{\pi} v=\{u \in V: \pi(u)<\pi(v)\}$ denotes the conditioning set of $v$. Conditional independence relations can simplify this factorization by reducing the conditioning sets. In probabilistic graphical models, Bayesian networks provide a compact representation of such factorizations through a directed graph whose structure encodes conditional independence relations.

\section{Relative rank networks}\label{sec:RN}
\emph{Relative rank networks} are graphical models for rankings obtained by combining RRVs with Bayesian network factorizations. In this section, we focus on RRVs defined by acyclic variable graphs, which break MEC and therefore induce partitions of the ranking space without empty events. This restriction avoids the structural zero constraints that arise when RRVs are subject to MEC and allows RRMs to be parameterized freely (see Appendix~\ref{sec:discussion.admissible}). Building on this representation, relative rank networks provide a principled way to represent complex ranking distributions using conditional independence assumptions.

Given an RRV defined by an acyclic variable graph, Eq.~\ref{eq:params.free} shows that an unrestricted RRM can require specifying an exponential number of parameters. Even for sparse variable graphs such as directed trees, this can lead to $\mathcal{O}(2^n)$ parameters. Moreover, probabilistic inference tasks, such as marginalization over subsets of RRVs, are computationally intractable in general. To address these challenges, we introduce graphical models that exploit conditional independence relations among RRVs.


\begin{definition}[Relative rank network]\label{def:RRN}
A \emph{relative rank network} (\emph{RRN}) is a triplet $(\V,\G,\T)$, where $\V$ is an acyclic variable graph defining the RRV $\X$ for $V$, $\G$ is a directed acyclic graph called the factorization graph, and $\T=(\theta_1,\dots,\theta_n)$ denotes the model parameters. The RRN $(\V,\G,\T)$ represents the RRM
\begin{equation}
h(\x;\G,\T)= \prod_{v \in V} h(x_v \mid \x_{P_v^\G}; \theta_v).
\end{equation}
\end{definition}

An RRN combines two complementary graphical structures. The variable graph $\V$ defines the RRVs and the partition of the ranking space they induce. The factorization graph $\G$ specifies how the joint distribution over these variables factorizes into conditional distributions. In this way, $\V$ determines the representation of rankings, while $\G$ determines the probabilistic structure of the model. We call $(\V,\G)$ the \emph{structure} of an RRN.

The number of parameters of an RRN depends on its structure. The graph $\V$ determines the size of the state space of each relative rank variable, while $\G$ determines the conditioning sets in the factorization. The total number of free parameters of an RRN with structure $(\V,\G)$ is
\begin{equation}
\sum_{v \in V} |P^{\V}_v|\prod_{u \in P_v^{\G}} \left(|P^{\V}_u| + 1\right).
\end{equation}
This flexibility allows the construction of RRNs with controlled complexity by restricting the maximum number of parents in both the variable graph $\V$ and the factorization graph $\G$. In particular, the number of free parameters is at most $nk^{s+1}$, where $k=\max_{v\in V}|\set{X}_v|$ and $s=\max_{v\in V}|P_v^{\G}|$; it grows linearly in $n$ when $k$ and $s$ are bounded.

For a fixed RRN structure, exact probabilistic inference over RRVs takes $\mathcal{O}(nk^{t+1})$ time~\citep{koller09}, where $k=\max_{v\in V}|\set{X}_v|$ is the maximum RRV cardinality and $t$ is the treewidth of the moralized factorization graph $\G$. For example, when $\G$ is a directed tree, $t=1$ and the cost reduces to $\mathcal{O}(nk^2)$.

RRNs can be connected to ranking models through a generalization of the pushforward mapping. Given a ranking model $p$, the pushforward under $(\V,\G)$ defines the parameters of an RRN by computing the conditional probabilities of the RRV realizations from $p$. This mapping allows us to focus on the structure $(\V,\G)$ independently of the specific parameter values.
\begin{definition}[Factorized pushforward]\label{def:factorized.push}
Let $\X$ be an RRV defined by an acyclic variable graph $\V$, and let $\G$ be a factorization graph. The \emph{factorized pushforward} of a ranking model $p \in \Delta(\sS_n^V)$ under $(\V,\G)$ is the function $\push_{\V,\G}: \Delta(\sS^V_n)\to\RRN$,
where $\RRN \subseteq \Delta(\sX)$ denotes the space of RRNs with structure $(\V,\G)$. The RRN $h=\push_{\V,\G}(p)$ is defined as
\[
h(\x)=\prod_{v\in V} h(x_v \mid \x_{P_v^\G}),
\]
where $h(x_v \mid \x_{P_v^\G})=
\frac{p(x_v,\x_{P_v^\G})}{p(\x_{P_v^\G})}$
if $p(\x_{P_v^\G})>0$, and $h(x_v \mid \x_{P_v^\G})=\frac{1}{|\set{X}_v|}$ otherwise, for all $v \in V$.
\end{definition}

The factorized pushforward mapping of a ranking model under $(\V,\G)$ produces an RRM that factorizes according to $\G$. Since in RRNs the RRV for $V$ breaks MEC, the factorized pushforward mapping yields admissible models, and any RRM can be specified without additional structural constraints. In Appendix~\ref{sec:discussion.cond.indep}, we discuss the admissibility of factorized models over RRVs subject to MEC and the structural zero constraints they impose.

For observed rankings and a fixed variable graph $\V$, learning $\G$ reduces to Bayesian-network structure learning over the categorical realizations $\X(\sigma)$, and the local distributions can be estimated from empirical frequencies. Appendix~\ref{sec:learning.RRN} outlines this procedure and the additional selection of $\V$ in Algorithm~\ref{alg:learning.RRN}.

The structure of an RRN admits a graphical interpretation in terms of conditional independence. The factorization graph $\G$ defines a Bayesian network over the RRV $\X=(X_1,\dots,X_n)$ defined by $\V$, and therefore encodes conditional independence relations among its components $X_v$ for $v \in V$. As in standard Bayesian networks, these relations can be derived from $\G$ using the d-separation criterion~\citep{pearl88}.

\begin{definition}[Independence model]\label{def:independence.model}
Let $(\V,\G)$ be the structure of an RRN. The \emph{independence model} induced by $(\V,\G)$ is the set of conditional independence relations
\begin{equation}
\dep(\V,\G) =
\{\;\X_A \perp \X_B \mid \X_C : A,B,C \subseteq V, \X_A \perp_{\G} \X_B \mid \X_C\},\nonumber
\end{equation}
where $\X$ is the RRV defined by $\V$, the subsets $A,B,C \subseteq V$ are pairwise disjoint with $A,B \neq \varnothing$, and the relation $\X_A \perp_{\G} \X_B \mid \X_C$ indicates that $A$ is d-separated from $B$ by $C$ in $\G$.
\end{definition}

Given the structure $(\V,\G)$, the induced independence model is a compositional graphoid satisfying the local Markov property~\citep{studeny2005independence}.

Any RRN with structure $(\V,\G)$ satisfies the conditional independence relations in $\dep(\V,\G)$.

\begin{lemma}\label{lemma:independence.model}
Let $h \in \RRN$ be an RRN with structure $(\V,\G)$, where $\X$ is the RRV defined by $\V$, and let $p \in \pull(h)$. Then $h$ and $p$ satisfy the conditional independence relations in $\dep(\V,\G)$, that is,
\[
\X_A \perp_{\G} \X_B \mid \X_C
\implies
\{\X_A \perp_h \X_B \mid \X_C,\quad \X_A \perp_p \X_B \mid \X_C\}.
\]
\end{lemma}

\section{Representing ranking models and independence assumptions}\label{sec:relation}
This section has two goals. First, we show how RRMs can be used to represent ranking models in general. Second, we show that two prominent families of ranking models based on alternative independence notions, namely repeated insertion and riffle shuffle models, can be represented within the RRN framework. In this way, the independence assumptions underlying both models correspond to specific independence models over RRVs.

\subsection{Ranking models based on RRMs}

An RRV $\X$ defined by an acyclic variable graph induces a partition of the ranking space, and RRMs for $\X$ represent probability distributions over this partition. A particularly useful situation occurs when an RRV induces a \emph{thin partition} of the ranking space, that is, a partition whose cells contain a single ranking. In this case, the mapping induced by the RRV from the ranking space $\sS^V_n$ to the state space $\sX$ is bijective, and therefore RRMs for $\X$ are ranking models.

For a general acyclic variable graph, an RRM specifies the probabilities of the events $\event{\x}$ but does not determine how probability is distributed among rankings within each event. A ranking model can be obtained by specifying a distribution within each event, as described in Appendix~\ref{sec:discussion.admissible}. For example, assigning equal probability to the rankings within each event yields $p(\sigma)=h(\X(\sigma))/|\event{\X(\sigma)}|$. The cardinality of the corresponding event can be obtained by enumerating its rankings using Algorithm~\ref{alg:event}.

The following result shows that an RRV defined by a complete acyclic variable graph induces a thin partition of the ranking space.

\begin{proposition}[Thin partition] \label{prop:complete.var.graph}
Let $\X$ be an RRV defined by $\V$. Then, $\X$ induces a \emph{thin partition} of $\sS^V_n$ if and only if $\V$ is a complete acyclic variable graph.
\end{proposition}

\begin{wrapfigure}{R}{0.50\textwidth}
\vspace{-2mm}
\centering

\begin{minipage}[c]{0.45\linewidth}
  \centering
  \includegraphics[height=2.6cm]{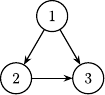}
\end{minipage}
\begin{minipage}[c]{0.45\linewidth}
  \centering
  \setlength{\tabcolsep}{3pt}
  \renewcommand{\arraystretch}{1.0}
  \setlength{\arrayrulewidth}{0.6pt}
  \begin{tabular}{cc|c}
    $\sigma$ & $\osigma$ & $\x$ \\ \hline
    \co{123}{1,2,3} & \co{123}{1,2,3} & \co{123}{1,2,3} \\
    \co{132}{1,3,2} & \co{132}{1,3,2} & \co{132}{1,2,2} \\
    \co{213}{2,1,3} & \co{213}{2,1,3} & \co{213}{1,1,3} \\
    \co{231}{2,3,1} & \co{231}{3,1,2} & \co{231}{1,2,1} \\
    \co{312}{3,1,2} & \co{312}{2,3,1} & \co{312}{1,1,2} \\
    \co{321}{3,2,1} & \co{321}{3,2,1} & \co{321}{1,1,1} \\
  \end{tabular}
\end{minipage}

\vspace{2mm}

\caption{%
The complete acyclic variable graph $\V=\{(1,2),(2,3),(1,3)\}$, where $\opi=[1,2,3]$ is its unique ancestral ordering. The RRV defined by $\V$ induces a thin partition of the ranking space.
}
\label{fig:example.complete}

\vspace{-4mm}
\end{wrapfigure}

Intuitively, a complete acyclic variable graph has a unique ancestral ordering $\opi$, and the reference set defining the RRV for $v=\opi(i)$ is $P_v=\{\opi(1),\dots,\opi(i-1)\}$ for $i=1,\dots,n$. Thus, a realization $\x$ encodes the position of each item with respect to all previously considered items in $\opi$, yielding a representation equivalent to the permutation itself. Figure~\ref{fig:example.complete} illustrates this correspondence for $n=3$, where the ancestral ordering is $\opi=[1,2,3]$.

An immediate consequence is that RRVs for $V$ defined by complete acyclic variable graphs represent bijective mappings between the state space and the ranking space, and therefore ranking models can be represented by RRMs. 
In the literature, RRVs providing bijective representations of permutations are known as \emph{inversion vectors}~\citep{malagon25bijective}, and they form the basis of repeated insertion models~\citep{doignon2004repeated}. According to Proposition~\ref{prop:complete.var.graph}, there are $n!$ different RRVs corresponding to inversion vectors.

In the remainder of this section, we use this representation to relate RRNs with two prominent families of ranking models based on independence notions tailored to permutations.

\subsection{Repeated insertion model}

Repeated insertion (RI) independence is a probabilistic independence notion defined with respect to a fixed ordering $\opi$ of the items, called the \emph{insertion ordering}~\citep{doignon2004repeated}. A ranking model $p$ satisfies RI independence for $\opi$~\citep{kenig2018rim} if it factorizes as
\begin{equation}
p(\sigma;\opi)= \prod_{i=1}^{n} p_i(f_{\sigma,\opi}(i)),\label{eq:RIM}
\end{equation}
for all $\sigma \in \sS^V_n$, where $p_i$ is the insertion distribution of the item $\opi(i)$ and \[f_{\sigma,\opi}(i)= 1 + \sum_{j=1}^{i-1} \mathbf{1}\{\sigma(\opi(j))< \sigma(\opi(i))\}.\]
The function $f_{\sigma,\opi}(i)$ determines the insertion position of the item $\opi(i)$ in the partial ranking $\sigma^{<_{\opi} i}$, where $<_{\opi} i=\{\opi(1),\dots,\opi(i-1)\}$. We call a \emph{repeated insertion model} (\emph{RIM}) for $\opi$ any ranking model that factorizes according to Eq.~\ref{eq:RIM}, and we denote \emph{the space of RIMs for $\opi$} by $\RIM$.

For a fixed insertion ordering and arbitrary categorical insertion distributions, a RIM has $n(n-1)/2$ free parameters.

From a generative perspective, a RIM for $\opi$ constructs rankings by sequentially inserting items according to the ordering $\opi$. Starting from the trivial ranking over $\{\opi(1)\}$, $(1)$, each item $\opi(i)$ is inserted into the current partial ranking at a position sampled from its insertion distribution $p_i$. After $n-1$ insertion steps, this procedure produces a full ranking for $V$. This generative process is a particular case of Algorithm~\ref{alg:event} in Appendix~\ref{sec:alg.event}.

\begin{theorem}\label{theo:RI.equivalence}
Let $\opi \in \sS^n_V$ be an insertion ordering. Let $\V^{RI}$ be the complete acyclic variable graph with ancestral ordering $\opi$, and let $\G^{RI}=\varnothing$ be the empty factorization graph. Let $\X$ be an RRV defined by $\V^{RI}$. Then,
\[
\RIM=\DeltaG{\G^{RI}}.
\]
\end{theorem}


As a consequence, the independence notion underlying RIMs can be expressed entirely in terms of conditional independence relations among RRVs.

\begin{corollary}\label{cor:RI.independence}
Let $\opi \in \sS^n_V$ be the insertion ordering, and let $\V^{RI}$ and $\G^{RI}$ be the graphs defined in Theorem~\ref{theo:RI.equivalence}. Then, RI independence for $\opi$ corresponds to the independence model $\dep(\V^{RI},\G^{RI})$.
\end{corollary}

In particular, since the factorization graph $\G^{RI}$ is empty, the independence model $\dep(\V^{RI},\G^{RI})$ contains all conditional independence relations $\X_A \perp \X_B \mid \X_C$ for disjoint non-empty subsets $A,B \subset V$ and $C\subseteq V\setminus (A \cup B)$.

An additional consequence of Theorem~\ref{theo:RI.equivalence} is that a RIM for $\opi$ can be constructed from any ranking model using the factorized pushforward mapping.

\begin{corollary}\label{cor:learn.RIM}
Let $\opi \in \sS^n_V$ be the insertion ordering, and let $\V^{RI}$ and $\G^{RI}$ be the graphs defined in Theorem~\ref{theo:RI.equivalence}. Let $p \in \Delta(\sS^V_n)$ be a ranking model. Then, $
h=\push_{\V^{RI},\G^{RI}}(p)$ is the RIM for $\opi$ that minimizes the KL divergence with respect to $p$, that is, $h=\argmin_{q \in \RIM} D_{KL}(p||q)$.
\end{corollary}

\subsection{Riffle shuffle model}
Riffle shuffle (RS) independence~\citep{huang2009riffled, huang2010learning, huang2012riffled} models rankings by independently ranking disjoint subsets of items and then interleaving the resulting partial rankings. The name is inspired by the riffle shuffle used in card games. Let $A=\{a_1,\dots,a_m\}$ and $B=\{b_1,\dots,b_{n-m}\}$ be a partition of $V$. A ranking model $p$ satisfies RS independence for $A$ and $B$ if it factorizes as
\begin{equation}
p(\sigma)=p_A(\sigma^A)p^B_A(s^B_A(\sigma))p_B(\sigma^B),\label{eq:RS}
\end{equation}
for every $\sigma \in \sS^V_n$, where $\sigma^A \in \sS^A_{|A|}$ and $\sigma^B \in \sS^B_{|B|}$ denote the (partial) rankings of the items in $A$ and $B$ induced by $\sigma$ (see Definition 12 in~\citep{huang2012uncovering}). The distributions $p_A \in \Delta(\sS^A_{|A|})$ and $p_B \in \Delta(\sS^B_{|B|})$ model the partial rankings within each subset of items. The function $s^B_A : \sS^V_n \to \binom{N}{|B|}$ determines how the positions of the items in $A$ and $B$ are interleaved in $\sigma$, and the distribution $p^B_A$ models the interleavings. For an interleaving $\bd{s}$, $\bd{s}(i)$ denotes its $i$-th smallest element. Given the interleaving $\bd{s}=s^B_A(\sigma)$ we have $\sigma(v)=\bd{s}(\sigma^B(v))$, for $v \in B$. We call a \emph{riffle shuffle model} (\emph{RSM}) for $A$ and $B$ any ranking model that factorizes according to Eq.~\ref{eq:RS}, and we denote \emph{the space of RSMs for $A$ and $B$} by $\RSM$.

An RSM for a fixed partition $A,B$, with arbitrary partial ranking and interleaving distributions, has $|A|!+|B|!+\binom{n}{|B|}-3$ free parameters: $|A|!-1$ for $p_A$, $|B|!-1$ for $p_B$, and $\binom{n}{|B|}-1$ for $p^B_A$.

From a generative perspective, an RSM constructs rankings in two stages. First, the items in $A$ and $B$ are ranked independently according to the distributions $p_A$ and $p_B$, producing the rankings $\sigma^A$ and $\sigma^B$. Then, the two rankings are interleaved by sampling an interleaving pattern from the distribution $p^B_A$, which determines the set of positions occupied by the items of each subset. The interleaving pattern is selected independently of $\sigma^A$ and $\sigma^B$.

We now show that RSMs can also be represented within the RRN framework by combining appropriate choices of variable graphs and a nontrivial factorization graph. In this representation, two RRN structures are considered, each defined by a different variable graph but sharing the same factorization graph. A ranking model that factorizes according to both structures corresponds to an RSM for $A$ and $B$.

\begin{theorem}\label{theo:RS.equivalence}
Let $A=\{a_1,\dots,a_m\}$ and $B=\{b_1,\dots,b_{n-m}\}$ form a partition of $V$. Let $\V^{RS}$ and $\W^{RS}$ be complete acyclic variable graphs with ancestral orderings $\opi=[a_1,\dots,a_m,b_1,\dots,b_{n-m}]$ and $\otau=[b_1,\dots,b_{n-m},a_1,\dots,a_m]$, respectively. Let $\X$ and $\Y$ be the RRVs defined by $\V^{RS}$ and $\W^{RS}$, respectively. Let $\G^{RS}=\G_A \cup \G_B$ be a directed acyclic graph, where $\G_A$ and $\G_B$ are complete subgraphs over the vertex sets $A$ and $B$, respectively. Then,
\[
\RSM = \Delta_{\G^{RS}}(\sX) \cap \Delta_{\G^{RS}}(\sY).
\]
\end{theorem}

RIMs correspond to a particularly simple instance of RRNs, characterized by a complete variable graph and an empty factorization graph. RSMs, in contrast, rely on a richer independence structure. In the RRN representation, this structure is captured by a nontrivial factorization graph over two RRVs for $V$ that provide two different relative-rank encodings of the ranking space, both inducing the same thin partition. 

According to Theorem~\ref{theo:RS.equivalence}, the space of RSMs is given by the intersection of the spaces of two RRNs. Thus, RS independence corresponds to the independence models specified by the structures of the two RRNs.

\begin{corollary}\label{cor:RS.independence}
Let $A$ and $B$ be a partition of $V$, and let $\V^{RS}$, $\W^{RS}$, and $\G^{RS}$ be the graphs defined in Theorem~\ref{theo:RS.equivalence}. Then, RS independence for $A$ and $B$ corresponds to $\dep(\V^{RS},\G^{RS}) \cup \dep(\W^{RS},\G^{RS})$.
\end{corollary}


Finally, the next result shows how to obtain an RSM for $A$ and $B$ from any ranking model using the factorized pushforward mapping.
\begin{proposition}\label{prop:learn.RSM}
Let $p$ be a ranking model, let $A$ and $B$ be a partition of $V$, and let $\V^{RS}$, $\W^{RS}$, and $\G^{RS}$ be the graphs defined in Theorem~\ref{theo:RS.equivalence}. Then, $h=\push_{\V^{RS},\G^{RS}}(\push_{\W^{RS},\G^{RS}}(p))$ is the RSM for $A$ and $B$ that minimizes the Kullback-Leibler divergence with respect to $p$.
\end{proposition}

Together, these results show that RIMs and RSMs can be represented within the RRN framework, each corresponding to a specific independence model over relative rank variables.

\section{Conclusions}\label{sec:conclusions}

This theoretical work introduces \emph{relative rank models} as a probabilistic framework for rankings based on structured coarsenings of the permutation space. By representing rankings through relative rank variables, the framework replaces the structure of permutations with categorical random variables that can break the mutual exclusivity constraints of rankings.

A central result of the paper is the characterization of acyclic variable graphs as the exact condition under which relative rank variables break mutual exclusivity constraints. When mutual exclusivity constraints are broken, admissibility imposes no further structural constraints on relative rank models, and relative rank models can be simplified by assuming conditional independence among relative rank variables.

Building on this characterization, we define \emph{relative rank networks}, which combine relative rank variables with Bayesian network factorizations. Relative rank networks provide a principled way to encode conditional independence relations among relative rank variables while retaining a clear semantics in terms of partitions of the ranking space. As a consequence, they enable the application of standard tools from probabilistic graphical models to ranking data.

We further show that well-known ranking models based on specialized independence notions, namely repeated insertion and riffle shuffle models, can be represented within the framework of relative rank networks. From this perspective, these models correspond to specific sets of conditional independences over relative rank variables rather than fundamentally different probabilistic constructions.

More generally, the results of this paper suggest a shift in how independence should be introduced in ranking models: instead of defining independence notions directly on permutations, one should first construct a structured coarsening that breaks mutual exclusivity constraints and then apply standard probabilistic independence relations. Relative rank networks provide a concrete and flexible realization of this principle.


\begingroup

\begin{ack}
The author thanks Ekhi\~ne Irurozki for helpful discussions. This work was supported by MCIN/AEI/10.13039/501100011033 under projects PLEC2024-011247 and CEX2021-001142-S, and by the Basque Government through the BERC 2022--2025 and Elkartek programs.
\end{ack}
\endgroup

\bibliography{relative.ranking}


\clearpage\appendix

\section{Examples of variable graphs and the induced partition}\label{app:variable.graphs}

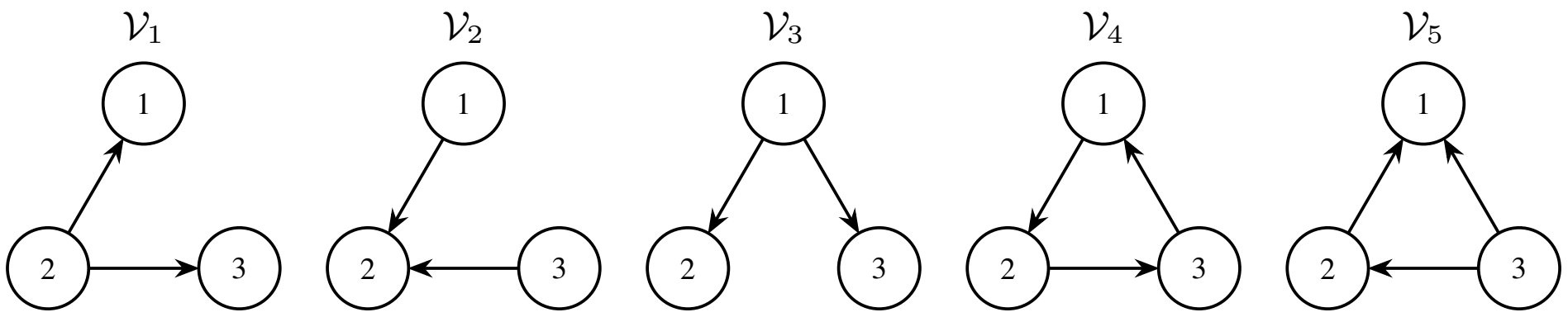
\begin{figure}[h]
	\centering
	\begin{tikzpicture}[
		scale=0.8,
		>={Stealth[length=2mm]}
		]
		
		\tikzset{mycircle/.style={circle, draw, fill=white, line width=0.7pt, minimum size=6.5mm, inner sep=0pt}}
		\tikzset{myarrow/.style={
				->,
				line width=0.7pt,
				shorten >=-0.2mm,
				shorten <=-0.2mm
		}}
		
		\def\xsep{3.2}
		\def\r{1.1}
		
		\foreach \i/\col/\name in {
			1/0/1,
			2/1/2,
			3/2/3,
			4/3/4,
			5/4/5
		}{
			\coordinate (Center\i) at (\col*\xsep, 0);
			
			\coordinate (A\i) at ($(Center\i)+(0,\r)$);
			\coordinate (B\i) at ($(Center\i)+(-0.87*\r,-0.5*\r)$);
			\coordinate (C\i) at ($(Center\i)+(0.87*\r,-0.5*\r)$);
			
			\node[mycircle] (N\i1) at (A\i) {\scriptsize 1};
			\node[mycircle] (N\i2) at (B\i) {\scriptsize 2};
			\node[mycircle] (N\i3) at (C\i) {\scriptsize 3};
			
			\node[anchor=south] at ($(A\i)+(0,0.45)$) {\small $\mathcal{V}_{\name}$};
		}
		
		
		\draw[myarrow] (N12) -- (N11);
		\draw[myarrow] (N12) -- (N13);
		
		\draw[myarrow] (N21) -- (N22);
		\draw[myarrow] (N23) -- (N22);
		
		\draw[myarrow] (N31) -- (N32);
		\draw[myarrow] (N31) -- (N33);
		
		\draw[myarrow] (N41) -- (N42);
		\draw[myarrow] (N42) -- (N43);
		\draw[myarrow] (N43) -- (N41);
		
		\draw[myarrow] (N52) -- (N51);
		\draw[myarrow] (N53) -- (N51);
		\draw[myarrow] (N53) -- (N52);
		
	\end{tikzpicture}
	
	\caption{
		Five variable graphs defined over items $V=\{1,2,3\}$.
	}
	\label{fig:variable.graph.reduced}
\end{figure}

Figure~\ref{fig:variable.graph.reduced} shows five variable graphs with the vertex set $V=\{1,2,3\}$. Each variable graph $\V$ defines a collection of reference sets $P_v$ for $v \in V$, where $P_v$ corresponds to the set of parents of $v$ in $\V$. The associated RRV $\X=(X_1,X_2,X_3)$ for $V$ is defined by
\[
X_v(\sigma):= r^v_{P_v}(\sigma), \qquad v \in V.
\]
The variable graphs and the reference sets that define the corresponding RRVs are:
\begin{itemize}
\item $\V_1=\{(2,1),(2,3)\}$: $P_1=\{2\}$, $P_2=\emptyset$, $P_3=\{2\}$.
\item $\V_2=\{(1,2),(3,2)\}$: $P_1=\emptyset$, $P_2=\{1,3\}$, $P_3=\emptyset$.
\item $\V_3=\{(1,2),(1,3)\}$: $P_1=\emptyset$, $P_2=\{1\}$, $P_3=\{1\}$.
\item $\V_4=\{(1,2),(2,3),(3,1)\}$: $P_1=\{3\}$, $P_2=\{1\}$, $P_3=\{2\}$.
\item $\V_5=\{(2,1),(3,1),(3,2)\}$: $P_1=\{2,3\}$, $P_2=\{3\}$, $P_3=\emptyset$.
\end{itemize}

Table~\ref{tab:rankings.and.variables.reduced} shows the relation between rankings $\sigma \in \sS^V_n$ (and orderings $\osigma \in \sS^n_V$) and the realizations $\x \in \sX$ of the RRVs defined by these graphs. For each variable graph, the table displays the cover $\event{\sX}$ of the ranking space $\sS^V_n$, where each realization $\x \in \sX$ and its corresponding event $\event{\x}$ is identified by a distinct color. For example, for $\V_1$ and realization $\x=(2,1,2)$, the event $\event{\x}=\{(2,1,3),(3,1,2)\}$ is highlighted in blue.

\begin{table}[h]
\centering

\begin{tabular}{cc|ccccc}
&& $\V_1$ & $\V_2$ & $\V_3$ & $\V_4$ & $\V_5$ \\
$\sigma$ & $\osigma$ & \multicolumn{5}{c}{$\x=(x_1,x_2,x_3)$} \\ \hline

\co{123}{1,2,3} & \co{123}{1,2,3} &
\co{123}{1,1,2} &
\co{123}{1,2,1} &
\co{123}{1,2,2} &
\co{123}{1,2,2} &
\co{123}{1,1,1} \\

\co{132}{1,3,2} & \co{132}{1,3,2} &
\co{132}{1,1,1} &
\co{132}{1,3,1} &
\co{123}{1,2,2} &
\co{132}{1,2,1} &
\co{132}{1,2,1} \\

\co{213}{2,1,3} & \co{213}{2,1,3} &
\co{213}{2,1,2} &
\co{213}{1,1,1} &
\co{213}{1,1,2} &
\co{213}{1,1,2} &
\co{213}{2,1,1} \\

\co{231}{2,3,1} & \co{231}{3,1,2} &
\co{132}{1,1,1} &
\co{132}{1,3,1} &
\co{231}{1,2,1} &
\co{231}{2,2,1} &
\co{231}{2,2,1} \\

\co{312}{3,1,2} & \co{312}{2,3,1} &
\co{213}{2,1,2} &
\co{213}{1,1,1} &
\co{312}{1,1,1} &
\co{312}{2,1,2} &
\co{312}{3,1,1} \\

\co{321}{3,2,1} & \co{321}{3,2,1} &
\co{321}{2,1,1} &
\co{123}{1,2,1} &
\co{312}{1,1,1} &
\co{321}{2,1,1} &
\co{321}{3,2,1} \\
\end{tabular}

\vspace{2mm}
\caption{
Relative rank variable realizations $\x=(x_1,x_2,x_3)$ for the rankings $\sigma\in\sS^V_n$ and orderings $\osigma\in\sS^n_V$, defined according to the five variable graphs shown in Figure~\ref{fig:variable.graph.reduced}. The graph $\V_4$ is cyclic and the realizations $\x \in \{(1,1,1),(2,2,2)\}$ do not appear in the table because they correspond to empty events.
}
\label{tab:rankings.and.variables.reduced}
\end{table}

The graph $\mathcal{V}_4$ is cyclic and therefore its associated RRV is subject to MEC, which implies that some realizations of the associated RRV correspond to empty events. In particular, the cyclic dependencies among the variables impose mutually inconsistent constraints on the relative ranks, so that some configurations cannot be satisfied by any ranking. As a consequence, the realizations $\x=(1,1,1)$ and $\x=(2,2,2)$ correspond to empty events, i.e., $\event{\x}=\varnothing$ (see Theorem~\ref{theo:break.MEC}). This prevents $\mathcal{V}_4$ from inducing a proper partition of the ranking space.

The graph $\mathcal{V}_5$ is a complete acyclic variable graph and therefore induces a thin partition of the ranking space. By Proposition~\ref{prop:complete.var.graph}, every realization $\x \in \sX$ corresponds to a singleton event, so that each ranking $\sigma \in \sS^V_n$ is uniquely identified by its image $\x=\X(\sigma)$. As a consequence, the RRVs defined by complete acyclic variable graphs define a bijection between the ranking space and their state space, and the RRMs are ranking models, $\Delta(\sX)=\Delta(\sS^V_n)$. In particular, any ranking model $p$ can be represented exactly as an RRM $h=\push(p)$, satisfying $p(\sigma)=h(\X(\sigma))$ for all $\sigma \in \sS^V_n$, and $p(\x)=\sum_{\sigma \in \event{\x}}p(\sigma)= h(\x)$ for all $\x \in \sX$.

\section{Generating the events}\label{sec:alg.event}
Thanks to the acyclicity of the variable graph, all the rankings belonging to a given event can be enumerated using Algorithm~\ref{alg:event}. The algorithm is an insertion-based procedure that follows an ancestral ordering of the variable graph to incrementally construct a set of partial rankings. At each step, a new item is inserted into each ranking from the previous iteration at positions determined by its associated relative rank.

\begin{algorithm}[H]
\caption{Event construction from an RRV realization}
\label{alg:event}

\textbf{Input:} A realization $\x$ of an RRV for $V$, $\X$, defined by an acyclic variable graph $\V$ with an ancestral ordering $\bar{\pi} \in \anc(\V)$

\textbf{Output:} The non-empty event associated to $\x$, $\event{\x}=\R_n$

\begin{algorithmic}[1]
\STATE Choose an ordering $\opi \in \anc(\V)$
\STATE Initialize $\R_1 \gets \{(1)\} = \sS^{\{\opi(1)\}}_1$

\FOR{$i = 2,\dots,n$}
    \STATE Initialize $\R_i \gets \varnothing$
    \STATE Let $v \gets \opi(i)$
    \FOR{each $\alpha \in \R_{i-1}$}
        \IF{$P^{\V}_v=\varnothing$}
            \STATE $I_v \gets \{1,\dots,i\}$
        \ELSIF{$x_v = 1$}
            \STATE $I_v \gets \{1,\dots,\alpha(\oalpha^{P_v^{\V}}(1))\}$
        \ELSIF{$x_v \in \{2,\dots,|P_v^{\V}|\}$}
            \STATE $I_v \gets 
            \big\{\alpha\big(\oalpha^{P_v^{\V}}(x_v-1)\big)+1,
            \dots,\alpha\big(\oalpha^{P_v^{\V}}(x_v)\big)\big\}$
        \ELSE 
            \STATE $I_v \gets \{\alpha(\oalpha^{P_v^{\V}}(|P_v^{\V}|))+1,\dots,i\}$
        \ENDIF
        \STATE $\R_i \gets \R_i \cup \{\ins(v, r, \alpha) : r \in I_v\}$
    \ENDFOR
\ENDFOR

\RETURN $\R_n$
\end{algorithmic}
\end{algorithm}

\begin{definition}[Insertion]\label{def:ins}
Let $U \subsetneq V$ with $m = |U|$, let $\alpha \in \sS^U_m$ be a (partial) ranking on $U$, let $v \in V \setminus U$, and let $r \in \{1,\dots,m+1\}$. The \emph{insertion operator}
\[
\ins(v,r,\alpha) = \beta \in \sS^{U \cup \{v\}}_{m+1}
\]
is defined by
\[
\beta(v) = r, \text{ and }
\beta(u) = \alpha(u) + \mathbf{1}\{\alpha(u) \ge r\} \text{ for } u \in U.
\]
\end{definition}

The resulting ranking on $U \cup \{v\}$, $\beta$, corresponds to the ordering
\[
\obeta = [\oalpha(1),\dots,\oalpha(r-1),v,\oalpha(r),\dots,\oalpha(m)].
\]

We consider the direct extensions of Definitions~\ref{def:relative.rank} and~\ref{def:RRV} to partial rankings containing the item and its reference set.

\begin{lemma}[Relative rank preservation]\label{lemma:RR.preservation}
Consider the set of items $U \subsetneq V$ with size $m=|U|$, the (partial) ranking $\alpha \in \sS^U_m$, the item $v \in V \setminus U$, the rank $r \in \{1,\dots,m+1\}$, and the ranking  $\beta = \ins(v,r,\alpha) \in \sS^{U \cup \{v\}}_{m+1}$. Then, for any $u \in U$ and any $P \subseteq U \setminus \{u\}$,
\[
r^u_P(\beta) = r^u_P(\alpha).
\]
\end{lemma}

\begin{proof}
Fix $u \in U$ and $P \subseteq U \setminus \{u\}$. Define the sets
\[
A = \{w \in P : \alpha(w) < r\},
\qquad
B = \{w \in P : r \le \alpha(w)\}.
\]
By Definition~\ref{def:ins}, we have:
\begin{itemize}
\item if $w \in A$, then $\beta(w) = \alpha(w)$,
\item if $w \in B$, then $\beta(w) = \alpha(w) + 1$.
\end{itemize}

We consider two cases for $\alpha(u)$.

\medskip
\noindent
i) $\alpha(u) < r$:

Then $\beta(u) = \alpha(u)$, and
\[
r^u_P(\beta)
=
1 + \sum_{w \in A} \mathbf{1}\{\beta(w) < \beta(u)\}
+
\sum_{w \in B} \mathbf{1}\{\beta(w) < \beta(u)\}.
\]
Substituting the definition of $\beta$,
\[
r^u_P(\beta)= 1+ 
\sum_{w \in A} \mathbf{1}\{\alpha(w) < \alpha(u)\}
+
\sum_{w \in B} \mathbf{1}\{\alpha(w)+1 < \alpha(u)\}.
\]
Since $\alpha(u) < r \le \alpha(w)$ for all $w \in B$, we have
\[
r^u_P(\beta)
= 1+ 
\sum_{w \in A} \mathbf{1}\{\alpha(w) < \alpha(u)\} + \sum_{w \in B} \mathbf{1}\{\alpha(w) < \alpha(u)\}
=
r^u_P(\alpha).
\]

\medskip
\noindent
ii) $\alpha(u) \ge r$:

Then $\beta(u) = \alpha(u) + 1$, and
\[
r^u_P(\beta)
=
1+ \sum_{w \in A} \mathbf{1}\{\beta(w) < \beta(u)\}
+
\sum_{w \in B} \mathbf{1}\{\beta(w) < \beta(u)\}.
\]
Substituting,
\[
r^u_P(\beta)= 1 +
\sum_{w \in A} \mathbf{1}\{\alpha(w) < \alpha(u)+1\}
+
\sum_{w \in B} \mathbf{1}\{\alpha(w)+1 < \alpha(u)+1\}.
\]
Since $\alpha(w) < r \leq \alpha(u)$ for all $w \in A$, we have
\[
r^u_P(\beta)
= 1+
\sum_{w \in A} \mathbf{1}\{\alpha(w) < \alpha(u)\} + \sum_{w \in B} \mathbf{1}\{\alpha(w) < \alpha(u)\}
=
r^u_P(\alpha).
\]
\end{proof}

The next result shows that for any $\x \in \sX$ Algorithm~\ref{alg:event} generates all the rankings in the event $\event{\x}$.
\begin{proposition}\label{prop:algorithm.event}
Let $\X$ be an RRV defined by an acyclic variable graph $\V$. Then, given a realization $\x \in \sX$, Algorithm \ref{alg:event} generates the rankings in the event $\event{\x} \neq \emptyset$. 
\end{proposition}
\begin{proof}
Let $\X$ be an RRV defined by $\V$ with the reference sets $P_v=P^{\V}_v$ for $v=1,\dots,n$. We have to prove that $\event{\x}=\R_n$ for arbitrary $\x \in \sX$. 

$\V$ is a DAG and thus there exists $\opi \in \anc(\V)$. Without loss of generality, assume that $\opi=[1,\dots,n]$. By definition of an ancestral ordering, the reference set $P_v \subseteq \{\opi(1),\dots,\opi(v-1)\}$, for $v=1,\dots,n$. 

We prove the result by induction over $i=1,\dots,n$: for $v=\opi(i)=i$, the event $\event{\x_{\{1,\dots,v\}}}=\bigcup_{\alpha \in \R_i}\event{\alpha}$, where $\event{\alpha}=\{\sigma: \sigma^{\{1,...,v\}}=\alpha\}$ and $|\R_i|\geq 1$.

\emph{Base case ($i=v=1$).}\\
The parents of $v$ in $\V$ are $P_1=\emptyset$. Then $\set{X}_v=\{1\}$ and thus $x_v=1$ with $\event{x_v}=\sS^V_n$. 

Using Algorithm~\ref{alg:event}, we have $\R_1=\{(1)\}=\sS^{\{1\}}_1$, and $\bigcup_{\alpha \in \R_1}\event{\alpha}=\event{(1)}=\sS^V_n=\event{x_v}.$

\emph{Inductive step.}\\
For $1\leq i<n$, let us assume that $\event{\x_{\{1,\dots,i\}}}= \bigcup_{\alpha \in \R_i} \event{\alpha}$ where $\R_i \subseteq \sS^{\{1,\dots,i\}}_i$ with $|\R_i| \geq 1$. Let $v=\opi(i+1)=i+1$. 

The ancestral ordering is $\opi=[1,\dots,n]$ and $P_v \subseteq \{1,\dots,i\}$. By Definition \ref{def:RRV} we have $\event{\x_{\{1,\dots,v\}}}= \event{x_v} \cap \event{\x_{\{1,\dots,i\}}}$ and thus
$$ \event{\x_{\{1,\dots,v\}}} = \bigcup_{\alpha \in \R_i} \event{\alpha} \cap \event{x_v}=\bigcup_{\alpha \in \R_i} \event{\alpha,x_v},$$
where $\event{x_v}=\{\sigma: r^v_{P_v}(\sigma)= x_v\}$ and $\event{\alpha,x_v}=\{\sigma: \sigma^{\{1,\dots,i\}}=\alpha, r^v_{P_v}(\sigma)= x_v\}$. To represent $\event{\x_{\{1,\dots,v\}}}$,  we have to construct the set of partial rankings $\{\beta \in \sS^{\{1,\dots,v\}}_v: \beta^{\{1,\dots,i\}}=\alpha,  r^v_{P_v}(\beta)= x_v\}$ for each partial ranking $\alpha \in \R_i$. By Definition~\ref{def:ins} and Lemma~\ref{lemma:RR.preservation} these partial rankings are obtained by inserting $v$ in $\alpha$ in the ranks $I_v=\{r \in \{1,\dots,i+1\}: \sum_{u \in P_v} \1{\alpha(u)<r} = x_v-1\}$:
$$\event{\alpha,x_v}= \bigcup_{r \in    I_v} \event{\ins(v,r,\alpha)}.$$

Note that for $r \in I_v$ and $\beta= \ins(v,r,\alpha)$ we have

$$X_v(\beta)=r^v_{P_v}(\beta)= 1 + \sum_{w \in P_v}\1{\beta(w)<\beta(v)}= 1 + \sum_{w \in P_v}\1{\beta(w)<r}= x_v.$$

Given $\alpha \in \sS^{\{1,\dots,i\}}_i$,  the set of insertion ranks $I_v$ satisfying the constraint $\sum_{u \in P_v} \1{\alpha(u)<r} = x_v-1$ is $\{1,\dots,i+1\}$ if $P_v=\varnothing$. Otherwise:
\begin{itemize}
    \item $I_v=\{1,\dots,\alpha(\oalpha^{P_v}(1))\},$ if $x_v=1$
    \item $I_v=\{\alpha(\oalpha^{P_v}(x_v-1))+1,\dots,\alpha(\oalpha^{P_v}(x_v))\},$ if $1<x_v\leq|P_v|$
    \item and $I_v=\{\alpha(\oalpha^{P_v}(|P_v|))+1,\dots,i+1\},$ if $x_v=|P_v|+1$.
\end{itemize}    
Finally, the set of insertion ranks satisfies $|I_v|\geq 1$ and thus $|\R_v|\geq 1$, for $v=2,\dots,n$. Thus, $\event{\x}= \R_n \neq \varnothing$.
\end{proof}

\section{Admissible models}\label{sec:discussion.admissible}
In this appendix we consider RRVs defined according to arbitrary variable graphs and we characterize admissible RRMs through the pushforward relation with ranking models. Recall that an RRM is admissible if and only if it is the pushforward of at least one ranking model. We first provide a constructive description of the pullback of an admissible RRM, and then derive a simple necessary and sufficient condition for admissibility based on the support of the model. This characterization shows that admissibility imposes a structural constraint on the model: An RRM must assign zero probability to all realizations associated with empty events.

\begin{lemma}[Pullback ranking models]\label{lemma:pullback.models}
Let $\X$ be an RRV for $V$ and $h$ be an RRM for $\X$, and let $p \in \Delta(\sS^V_n)$ be a ranking model. Then, $p \in \pull(h)$ if and only if for all $\x \in \sX$ with $h(\x)>0$ there exists $p_{\x} \in \Delta(\event{\x})$ such that
\begin{equation}
p(\sigma)= h(\x)p_{\x}(\sigma),\label{eq:pullback}
\end{equation}
for $\sigma \in \event{\x}$.
\end{lemma}
\begin{proof}
\begingroup
By Definition~\ref{def:push.pull}, $p \in \pull(h)$ if and only if $h=\push_{\X}(p)$.

i) Suppose that for every $\x$ with $h(\x)>0$ there exists $p_{\x} \in \Delta(\event{\x})$ satisfying Eq.~\ref{eq:pullback}. Summing within each such event gives
\[
\sum_{\sigma\in\event{\x}}p(\sigma)
=h(\x)\sum_{\sigma\in\event{\x}}p_{\x}(\sigma)
=h(\x).
\]
These events are disjoint, and their total probability under $p$ is $\sum_{\x:h(\x)>0}h(\x)=1$. Since $p$ is a probability distribution, every remaining event has probability zero. Hence $\push_{\X}(p)=h$.

ii) Suppose that $h=\push_{\X}(p)$. For each $\x$ with $h(\x)>0$, define $p_{\x}(\sigma)=p(\sigma)/h(\x)$ for $\sigma\in\event{\x}$. This is a probability distribution on $\event{\x}$, because its values are nonnegative and sum to one, and it satisfies Eq.~\ref{eq:pullback}.

If $h(\x)=0$, then $p(\sigma)=0$ for every $\sigma\in\event{\x}$. For $\event{\x}\neq\varnothing$, any choice of $p_{\x}\in\Delta(\event{\x})$ satisfies Eq.~\ref{eq:pullback}; if $\event{\x}=\varnothing$, no $p_{\x}$ needs to be defined.
\endgroup
\end{proof}
Using Eq.~\ref{eq:pullback}, we can generate ranking models in the pullback of any admissible RRM by choosing a distribution $p_{\x}$ within each event with $h(\x)>0$. A particularly simple choice that does not require extra parameters assigns equal probability to all rankings within each such event.

The next result provides a simple rule to characterize the admissible models.
\begin{proposition}[Structural zeroes]\label{prop:admisible.pushforward}
Let $\X$ be an RRV for $V$. An RRM $h \in \Delta(\sX)$ is admissible if and only if $h(\x)=0$ for all $\x \in \sX$ with $\event{\x}= \varnothing$.
\end{proposition}
\begin{proof}
The proof has two parts. 

i) If $h$ is admissible then $h(\x)=0$ for all $\x \in \sX$ with $\event{\x}=\varnothing$: 

If $h \in \Delta(\sX)$ is admissible by Definition~\ref{def:admissible} then there exists $p \in \Delta(\sS^V_n)$ such that $h(\x)=p(\x):=\sum_{\sigma \in \event{\x}} p(\sigma)$ for all $\x \in \sX$. Therefore, for $\x^{\varnothing} \in \sX$ with $\event{\x^{\varnothing}}= \varnothing$ we have $h(\x^{\varnothing})=p(\x^{\varnothing})=0$.

ii) If $h(\x)=0$ for all $\x \in \sX$ with $\event{\x}=\varnothing$  then $h$ is admissible: 

By Definition~\ref{def:admissible}, $h$ is admissible if there exists $p \in \pull(h)$. According to Lemma~\ref{lemma:pullback.models} $p \in \pull(h)$ if and only if
\[p(\sigma)= h(\x)p_{\x}(\sigma),\] 
where $p_{\x} \in \Delta(\event{\x})$ for $\x \in \sX$ with $h(\x)>0$. 

We obtain a $p \in \pull(h)$ by construction. For $\x \in \sX$ with $h(\x)>0$, define the uniform distribution $p_{\x} \in \Delta(\event{\x})$ as $p_{\x}(\sigma)= 1/|\event{\x}|$ if $\sigma \in \event{\x}$ and $p_{\x}(\sigma)=0$ otherwise. Set $p(\sigma)=h(\X(\sigma))p_{\X(\sigma)}(\sigma)$ when $h(\X(\sigma))>0$, and $p(\sigma)=0$ otherwise. Then $p \in \Delta(\sS^V_n)$ because $p(\sigma)\geq 0$ and 
\[
\sum_{\sigma \in \sS^V_n}p(\sigma)=\sum_{\x: \event{\x} \neq \varnothing} \sum_{\sigma \in \event{\x}} p(\sigma)= \sum_{\substack{
  \x: \event{\x} \neq \varnothing, \\
  h(\x) > 0
}} h(\x) \sum_{\sigma \in \event{\x}} p_{\x}(\sigma)
= \sum_{\substack{
  \x: \event{\x} \neq \varnothing, \\
  h(\x) > 0
}} h(\x)= 1,
\]
where the last equality holds because $h(\x)= 0$ for all $\x$ with $\event{\x}=\varnothing$. Therefore $p \in \pull(h)$ and $\pull(h) \neq \varnothing$.
\end{proof}
All RRMs defined over variables that break MEC are admissible, since every realization corresponds to a non-empty event. In contrast, when RRVs are subject to MEC, admissibility requires assigning zero probability to all realizations associated with empty events.

\section{Admissible models and conditional independence}\label{sec:discussion.cond.indep}

When an RRV is defined by an acyclic variable graph, it breaks MEC, and all realizations correspond to non-empty events. In this case, admissibility does not require additional constraints, and conditional independence relations can be imposed without requiring structural zeros in the associated RRMs.

In contrast, when an RRV is defined by a cyclic variable graph, it is subject to MEC, and some realizations correspond to empty events. In this setting, certain conditional independence relations require imposing structural zeros on the associated RRMs in order to ensure admissibility. Moreover, these constraints propagate to ranking models in the pullback, potentially requiring structural zeros over an exponential number of rankings.

The following result illustrates this phenomenon.
\begin{theorem}
Let $A,B,C$ be three pairwise disjoint nonempty subsets of $V$. Let $\X$ be an RRV for $V$ with $\x^\varnothing \in \sX$ so that $\event{\x^\varnothing_{A,B,C}}=\varnothing$. 

A ranking model $p \in \Delta(\sS^V_n)$ satisfies $\X_A \perp_p \X_B \mid \X_C$ only if
\[
p(\x^\varnothing_C)=0 \quad \text{or} \quad p(\x^\varnothing_{A,C}) = 0 \quad \text{or} \quad p(\x^\varnothing_{B,C}) = 0.
\]
\end{theorem}

\begin{proof}

We argue by contradiction. Suppose that there exists $p \in \Delta(\sS^V_n)$, with $p(\x^\varnothing_C)>0$, $p(\x^\varnothing_{A,C})>0$ and $p(\x^\varnothing_{B,C})>0$: 

The event associated to the realization of $\x^\varnothing_{A,B,C}$ is empty, $\event{\x^\varnothing_{A,B,C}}=\varnothing$, and thus 
\[
p(\x^\varnothing_{A,B,C})= \sum_{\sigma \in \event{\x^\varnothing_{A,B,C}}} p(\sigma)= 0.
\]

In addition, by Definition~\ref{def:conditional.independence}, the conditional independence $\X_A \perp_p \X_B \mid \X_C$ implies that $p(\x_{A,B}|\x_{C})= p(\x_{A}|\x_{C})p(\x_{B}|\x_{C})$ for all $\x_{A,B,C} \in \sX_{A,B,C}$ with $p(\x_C)>0$.

Since $p(\x^\varnothing_C)>0$, we have
\[
p(\x^\varnothing_{A,B,C}) = \frac{p(\x^\varnothing_{A,C})\, p(\x^\varnothing_{B,C})}{p(\x^\varnothing_{C})}>0.
\]
This contradicts $p(\x^\varnothing_{A,B,C}) = 0$.
\end{proof}
Note that $\event{\x^{\varnothing}_{A,C}} \subseteq \event{\x^\varnothing_{C}}$ and $\event{\x^\varnothing_{B,C}}  \subseteq \event{\x^\varnothing_{C}}$. This result implies that $p$ must assign zero probability to all rankings in $\event{\x^\varnothing_{A,C}}$ or in $\event{\x^\varnothing_{B,C}}$. The number of such rankings can be exponential in the size of $V$, making these constraints highly restrictive. 

This phenomenon extends the notion of probabilistic independence for rankings introduced in~\citep{huang2009exploiting}. In that framework, independence between subsets of items is defined directly in terms of rankings, leading to strong structural constraints. In comparison, RRVs provide a more flexible representation, but when the MEC are present, enforcing conditional independence may still require a large number of structural zeros.

In contrast, for RRVs that break MEC, all realizations correspond to non-empty events, admissibility imposes no structural restrictions, and conditional independence assumptions lead to valid factorizations of RRMs. Consequently, acyclic RRVs provide the appropriate foundation for developing graphical models over rankings. The separation between the construction of the partition and the imposition of independence assumptions is a key advantage of RRMs.

\begingroup

\section{Learning relative rank networks}\label{sec:learning.RRN}

Let $\mathcal{D}=(\sigma^{(1)},\dots,\sigma^{(M)})$ be a sample of $M$ rankings, and let $\hat p$ denote its empirical distribution. For a fixed acyclic variable graph $\V$, transform the sample into the categorical data $\mathcal{D}_{\X}=(\X(\sigma^{(1)}),\dots,\X(\sigma^{(M)}))$. A factorization graph $\G$ can then be learned using standard score-based, constraint-based, or hybrid Bayesian-network structure-learning methods~\citep{koller09}. For fixed $(\V,\G)$, estimating the local conditional distributions by empirical frequencies gives $h=\push_{\V,\G}(\hat p)$, with the convention for unobserved parent configurations specified in Definition~\ref{def:factorized.push}.

Selecting $\V$ requires comparing different coarsenings of the ranking space. The selection criterion should account for the ranking information retained and the number of model parameters. Raw RRM likelihoods alone do not provide this comparison: an empty variable graph maps every observation to the same realization, which can be assigned probability one, giving log-likelihood zero regardless of the sample. Algorithm~\ref{alg:learning.RRN} therefore takes a model-selection criterion as an input. Defining and evaluating such a criterion remains future work.

\begin{algorithm}[H]

\caption{A general procedure for learning an RRN}
\label{alg:learning.RRN}
\textbf{Input:} Ranking sample $\mathcal{D}$; a finite non-empty collection $\mathfrak{V}$ of candidate acyclic variable graphs; a categorical Bayesian-network structure learner; a model-selection criterion, with higher values preferred

\textbf{Output:} A selected RRN $(\V^*,\G^*,\T^*)$

\begin{algorithmic}[1]
\FOR{each $\V\in\mathfrak{V}$}
    \STATE Define $\X$ from $\V$ and form $\mathcal{D}_{\X}=(\X(\sigma^{(j)}))_{j=1}^M$
    \STATE Learn a factorization graph $\G$ from $\mathcal{D}_{\X}$ using the supplied structure learner
    \STATE Estimate $\T$ from empirical conditional frequencies, as in Definition~\ref{def:factorized.push}
    \STATE Evaluate $(\V,\G,\T)$ using the supplied model-selection criterion
\ENDFOR
\STATE Select a highest-scoring RRN $(\V^*,\G^*,\T^*)$ among the fitted candidates
\RETURN $(\V^*,\G^*,\T^*)$
\end{algorithmic}
\end{algorithm}

The candidate variable graphs may be restricted by their maximum number of parents to control RRV cardinalities. Likewise, the learner for $\G$ may impose indegree or treewidth constraints to control parameter counts or inference cost. More elaborate searches could alternate local changes in $\V$ and $\G$, recomputing the categorical data when $\V$ changes. Developing and empirically evaluating these searches remains future work.
\endgroup

\section{Proofs of Theoretical Results}
\label{app:proofs}
This appendix provides the proofs of the theoretical results in the main text.

\subsection{{Proof of Theorem~\ref{theo:break.MEC}} (Breaking MEC)}\label{proof:break.MEC}
\begin{proof}
i) If $\X$ breaks MEC, then $\V$ is acyclic (by contradiction):

Assume that $\X$ is defined by a cyclic graph $\V$ and breaks MEC. Suppose for contradiction that $\V$ has a directed cycle
$$u_1,...,u_l,u_1$$ 
for some length $1 \leq l \leq n$. Let $U=\{u_1,\dots,u_l\}$ be the set of items in the cycle. Now define a realization $\x_U^\varnothing=(1,1,...,1)$ that forces a strict cyclic pairwise preference. Since $u_i \in P_{u_{i+1}}$ for $i=1,\dots,l-1$ and $u_l \in P_{u_1}$, $\x_U^\varnothing$ implies
$$\sigma(u_1)>\sigma(u_2)>\dots>\sigma(u_l)>\sigma(u_1)$$
for any $\sigma \in \event{\x_U^\varnothing}$, which is impossible. So $\event{\x_U^{\varnothing}}=\varnothing$.

Now take a realization $\x=(\x_{V\setminus U},\x^\varnothing_U)$. We have
$$\event{\x} = \event{\x_{V\setminus U}} \cap \event{\x_U^\varnothing}\subseteq \event{\x_U^\varnothing}= \varnothing,$$
which contradicts the assumption that $\X$ breaks MEC. Therefore, $\V$ must be acyclic.

ii) If $\V$ is acyclic, then $\X$ breaks MEC:

If $\X$ is defined by an acyclic variable graph, then by Proposition~\ref{prop:algorithm.event} for any $\x \in \sX$ we have $|\event{\x}|\geq 1$, and thus $\X$ breaks MEC.
\end{proof}

\subsection{Proof of Lemma~\ref{lemma:independence.model}}
\begin{proof}
By Theorem~\ref{theo:break.MEC}, an RRV $\X$ defined by an acyclic variable graph $\V$ breaks MEC. Then, by Proposition~\ref{prop:admisible.pushforward}, any RRN $h$ for $\X$ is admissible, that is, $\pull(h)\neq \varnothing$. 

By Definition~\ref{def:RRN}, $h$ factorizes according to $\G$.

On one hand, according to Theorem 3.3 in \citet{koller09}, if a distribution $h$ factorizes according to $\G$ and the sets $A$ and $B$ are d-separated by $C$ in $\G$, then $\X_A \perp_h \X_B | \X_C$. On the other hand, $p \in \pull(h)$, and thus $h=\push_{\X}(p)$, and by Definition~\ref{def:conditional.independence} $\X_A \perp_p \X_B | \X_C$.
\end{proof}


\subsection{Proof of Proposition~\ref{prop:complete.var.graph} (Thin partition)}
\begin{proof}

i) If the partition induced by $\X$ is thin, then $\V$ is acyclic complete (by contradiction):

First, assume that $\V$ is cyclic. Then, according to Theorem~\ref{theo:break.MEC}, there exists $\x \in \sX$ with $\event{\x}=\emptyset$, and thus $\event{\sX}$ is not thin. Therefore $\V$ is acyclic.

Next, assume that $\V$ is not complete. Then the size of the state space of $\X$ is smaller than $n!$. The size of the ranking space $\sS^V_n$ is $n!$, and thus there is at least one realization $\x$ with an associated event with size larger than $1$. Then the partition $\event{\sX}$ is not thin. Therefore $\V$ is complete.

ii) If $\V$ is acyclic complete then the partition induced by $\X$ is thin:

The RRV $\X$ defined by an acyclic $\V$ breaks MEC, and thus $\event{\sX}$ is a partition of $\sS^V_n$. Besides, the size of the state space of $\X$ when $\V$ is complete is $n!$. The only partition of $\sS^V_n$ with $n!$ non-empty subsets is a thin partition.
\end{proof}


\subsection{Proof of Theorem~\ref{theo:RI.equivalence}}

\begin{proof}
Without loss of generality, we assume that $\opi=[1,\dots,n]$.

$\X$ is defined by $\V^{RI}$ that is a complete DAG, and by Proposition~\ref{prop:complete.var.graph}, $\X$ induces a thin partition. Then, for simplicity, for any $\x \in \sX$ with $\event{\x}=\{\sigma\}$ we write $h(\sigma)=h(\x)$, and $p(\x)=p(\sigma)$. 

i) $\Delta_{\G^{RI}}(\sX) \subseteq \RIM$:

Let $h \in \Delta_{\G^{RI}}(\sX)$. Then, by Definition~\ref{def:RRN}
\[h(\x)=\prod_{v=1}^{n} h(x_v),\]
for all $\x \in \sX$, where $X_v(\sigma):= r^v_{P_v}(\sigma)$ for $v\in V$. $\V^{RI}$ is a complete DAG with the ancestral ordering $\opi$ and thus $X_v(\sigma):= r^v_{\{1,\dots,v-1\}}(\sigma)=f_{\sigma,\opi}(v)$, for $v\in V$. Therefore, by Eq.~\ref{eq:RIM}, $h$ is a RIM for $\opi$, and $\Delta_{\G^{RI}}(\sX) \subseteq \RIM$. 

ii) $\RIM \subseteq \Delta_{\G^{RI}}(\sX)$:

We prove it by showing that, for $p \in \RIM$, the RRN $h=\push_{\V^{RI},\G^{RI}}(p)$ is equivalent to $p$, i.e., $p(\sigma)=h(\X(\sigma))$ for all $\sigma \in \sS^V_n$.

Let $p \in \RIM$ and $h=\push_{\V^{RI},\G^{RI}}(p)$. Then by Definition~\ref{def:factorized.push} and (i)
\[h(\x)=\prod_{v=1}^{n} h(x_v)=\prod_{v=1}^{n} p(x_v)=\prod_{i=1}^{n} p_i(f_{\sigma,\opi}(i))=p(\sigma),\]
where $\event{\x}=\{\sigma\}$, for all $\x \in \sX$.
\end{proof}

\subsection{Proof of Corollary~\ref{cor:RI.independence}}
\begin{proof}
Let $\X$ be an RRV defined by $\V^{RI}$. By Theorem~\ref{theo:RI.equivalence} we have $\RIM = \Delta_{\G^{RI}}(\sX)$. Thus, by Lemma~\ref{lemma:independence.model}, RI independence is the independence model $\dep(\V^{RI},\G^{RI})$.
\end{proof}

\subsection{Proof of Corollary~\ref{cor:learn.RIM}}
To prove this result we use the notation introduced in the proof of Theorem~\ref{theo:RI.equivalence}.

\begin{proof}
Let $\X$ be an RRV defined by $\V^{RI}$ and let $h=\push_{\V^{RI},\G^{RI}}(p)$. By Definition~\ref{def:factorized.push}, we have $h \in \Delta_{\G^{RI}}(\sX)$, where $h(\x)=\prod_{v \in V}p(x_v)$, for all $\x \in \sX$. By Theorem~\ref{theo:RI.equivalence}, $h \in \RIM$.

The Kullback-Leibler divergence between $p$ and $g$ is:
\[D_{KL}(p||g)= \sum_{\sigma \in \sS^V_n} p(\sigma) \ln \frac{p(\sigma)}{g(\sigma)}.\]

We have that the minimization of the Kullback-Leibler divergence is equivalent to the next problems:
\begin{align}
 \argmin_{g \in \RIM}D_{KL}(p||g)
&= \argmax_{g \in \RIM} \sum_{\sigma \in \sS^V_n} p(\sigma) \ln g(\sigma)\nonumber\\
&=\argmax_{g \in \Delta_{\G^{RI}}(\sX)} \sum_{\x \in \sX} p(\x) \ln g(\x)\nonumber\\
&= \argmax_{g \in \Delta_{\G^{RI}}(\sX)} \sum_{\x \in \sX} p(\x) \ln \prod_{v \in V} g(x_v)\nonumber\\
&= \argmax_{g \in \Delta_{\G^{RI}}(\sX)} \sum_{v\in V} \sum_{x_v \in \mathcal{X}_v} p(x_v) \ln g(x_v).\nonumber
\end{align}

The first equality follows from the definition of KL divergence, since $\sum_{\sigma}p(\sigma)\ln p(\sigma)$ does not depend on $g$. The second follows from Theorem~\ref{theo:RI.equivalence} and the bijection between rankings and RRV realizations established by Proposition~\ref{prop:complete.var.graph}. The third follows from Definition~\ref{def:RRN}, since $\G^{RI}$ is empty. The fourth follows by expanding the logarithm of the product and marginalizing over the remaining variables.

Since the marginal distributions $g(x_v)$ can be chosen independently, the last objective can be maximized separately for each $v\in V$.
Therefore, a RIM that minimizes the Kullback-Leibler divergence with respect to $p$ is
\[g^*(\x)=\prod_{v \in V}p(x_v)=h(\x),\]
for all $\x \in \sX$.
\end{proof}

\subsection{Proof of Theorem~\ref{theo:RS.equivalence}}
To prove Theorem~\ref{theo:RS.equivalence}, we first introduce a result that relates RRVs with relative rankings when the variable graph is a complete acyclic graph.

Let $A$ and $B$ be two disjoint non-empty subsets. The relative ranking of $A$ with respect to $B$ under $\sigma$, $\sigma^A_B \in \sS^A_{|A|+|B|}$, is the injective function from $A$ to $\{1,\dots,|A|+|B|\}$ mapping each item to its rank in the partial ranking $\sigma^{A,B}$, that is, $\sigma^A_B(a)=\sigma^{A,B}(a)$ for $a \in A$. Here, $\sS^A_{|A|+|B|}$ denotes the space of these injective functions.

\begin{lemma}[Partial and relative rankings]\label{lemma:rel.rankings}
    Let $\V$ be a complete acyclic variable graph with ancestral ordering $\opi=[a_1,\dots,a_{m},b_1,\dots,b_{n-m}]$, let $A=\{a_1,...,a_{m}\}$ and $B=\{b_1,\dots,b_{n-m}\}$ be a partition of $V$, and let $\X$ be defined according to $\V$. Then 
    \begin{itemize}
    \item $\X_A$ induces a thin partition of the (partial) ranking space of $A$, $\sS^A_{m}$, and
    \item $\X_B$ induces a thin partition of the relative ranking space of $B$ with respect to $A$, $\sS^B_n$.
    \end{itemize}
\end{lemma}
\begin{proof}
i) $\X_A$ induces a thin partition of $\sS^A_{m}$:

The definition of $\X_A$ is equivalent to an RRV for $A$ defined by the complete acyclic variable graph $\V^{A}$ with ancestral ordering $\opi^A=[a_1,\dots,a_m]$. Thus by Proposition~\ref{prop:complete.var.graph}, $\X_A$ induces a thin partition of $\sS^A_{|A|}$.

ii) $\X_B$ induces a thin partition of $\sS^B_n$

\begingroup
For $j=1,\dots,n-m$, the reference set of $b_j$ is $A\cup\{b_1,\dots,b_{j-1}\}$. Given $\x_B$, start from any partial ranking $\alpha\in\sS^A_m$ and successively insert each $b_j$ at rank $x_{b_j}$ using the insertion operator of Definition~\ref{def:ins}. Each insertion rank is uniquely specified because the reference set contains all previously inserted items. Changing $\alpha$ changes only the partial ordering of the items in $A$, leaving the insertion ranks and the final ranks of the items in $B$ unchanged. Thus $\x_B$ uniquely determines an injective function from $B$ to $N$, independently of $\alpha$.

Conversely, given such an injective function, choose a ranking of $V$ in which the items in $B$ have the specified ranks and the partial ranking of $A$ is $\alpha$. Remove $b_{n-m},\dots,b_1$ in this order, recording the rank of each item immediately before removal. These ranks recover $x_{b_{n-m}},\dots,x_{b_1}$, respectively, independently of $\alpha$. Thus the two constructions are inverse, establishing a bijection between $\sX_B$ and $\sS^B_n$. In particular, for $\x=\X(\sigma)$,
\[
\event{\x_B}=\{\rho\in\sS^V_n:\rho^B_A=\sigma^B_A\},
\]
which proves the claim.
\endgroup
\end{proof}

Now, we proceed with the proof of Theorem~\ref{theo:RS.equivalence}.
\begin{proof}
We prove that $\RSM= \Delta_{\G^{RS}}(\sX) \cap \Delta_{\G^{RS}}(\sY) $, where $\X$ and $\Y$ are defined by $\V^{RS}$ and $\W^{RS}$, respectively, by showing that  $h \in \Delta_{\G^{RS}}(\sX) \cap \Delta_{\G^{RS}}(\sY)$ if and only if $h \in \RSM$.

The RRVs $\X$ and $\Y$ for $V$ are defined by the complete acyclic graphs $\V^{RS}$ and $\W^{RS}$, respectively. By Proposition~\ref{prop:complete.var.graph}, $\X$ and $\Y$ induce thin partitions. Then, for $h \in\Delta(\sX)$ and $\x \in \sX$ with $\event{\x}=\{\sigma\}$, we write $h(\sigma)=h(\x)$ and $p(\x)=p(\sigma)$ for simplicity. Moreover, for any $p \in \Delta(\sS^V_n)$, partial ranking $\sigma^A \in \sS^A_m$, and interleaving $\bd{s}^B_A \in \binom{N}{|B|}$, we denote $p(\sigma^A)=\sum_{\rho \in \event{\sigma^A}}p(\rho)$ and $p(\bd{s}^B_A)=\sum_{\rho \in \event{\bd{s}^B_A}}p(\rho)$, where $\event{\sigma^A}=\{\rho: \rho^A=\sigma^A\}$ and $\event{\bd{s}^B_A}= \{\rho \in \sS^V_n: s^B_A(\rho)=\bd{s}^B_A\}$.

An RRM $h \in \Delta_{\G^{RS}}(\sX)$ if and only if
\[h(\sigma)= h(\sigma^A)h(\sigma^B_A)=h(\sigma^A) h(s^B_A(\sigma),\sigma^B),\]
for all $\sigma \in \sS^V_n$. The first equality holds by Lemma~\ref{lemma:rel.rankings}. The second equality holds because for $v \in B$
\[
\sigma^B_A(v)=\bd{s}^B_A(\sigma^B(v)),
\] 
where $\bd{s}^B_A= s^B_A(\sigma)$, for $\sigma \in \sS^V_n$.

An RRM $h\in \Delta_{\G^{RS}}(\sY)$ if and only if
\[h(\sigma)= h(\sigma^A_B)h(\sigma^B)=h(\sigma^A, s^A_B(\sigma)) h(\sigma^B),\]
for all $\sigma \in \sS^V_n$. In addition, $s^A_B(\sigma)$ uniquely determines $s^B_A(\sigma)$, for $\sigma \in \sS^V_n$. Thus, we have that $h\in \Delta_{\G^{RS}}(\sY)$ if and only if
\[
h(\sigma)=h(\sigma^A,s^B_A(\sigma))h(\sigma^B),
\]
for $\sigma \in \sS^V_n$.

Then, $h \in \Delta_{\G^{RS}}(\sX) \cap \Delta_{\G^{RS}}(\sY)$ if and only if
\[
h(\sigma)=h(\sigma^A)h(s^B_A(\sigma))h(\sigma^B),
\]
for $\sigma \in \sS^V_n$. 

By Eq.~\ref{eq:RS}, $h \in \RSM$ if and only if
\[h(\sigma)=h(\sigma^A)h(s^B_A(\sigma))h(\sigma^B),\]
for all $\sigma \in \sS^V_n$. And therefore, $h \in \Delta_{\G^{RS}}(\sX) \cap \Delta_{\G^{RS}}(\sY)$ if and only if $h \in \RSM$.
\end{proof}

\subsection{Proof of Corollary~\ref{cor:RS.independence}}
\begin{proof}
Let $\X$ and $\Y$ be the RRVs defined by $\V^{RS}$ and $\W^{RS}$. By Theorem~\ref{theo:RS.equivalence} we have $\RSM \subseteq \Delta_{\G^{RS}}(\sX)$ and $\RSM \subseteq \Delta_{\G^{RS}}(\sY)$. Thus, by Lemma~\ref{lemma:independence.model}, RS independence is given by the union of independence models $\dep(\V^{RS},\G^{RS})$ and $\dep(\W^{RS},\G^{RS})$.
\end{proof}

\subsection{Proof of Proposition~\ref{prop:learn.RSM}}
In this proof we reuse the notation in the proof of Theorem~\ref{theo:RS.equivalence}.

\begin{proof}
Let $g=\push_{\W^{RS},\G^{RS}}(p) \in \Delta_{\G^{RS}}(\sY)$, and $h=\push_{\V^{RS},\G^{RS}}(g) \in \Delta_{\G^{RS}}(\sX)$.

i) $h \in \RSM$:

First, by Definition~\ref{def:factorized.push}, we have
\begin{align*}
g(\sigma)=&p(\sigma^A_B)p(\sigma^B)\\
=&p(\sigma^A,\bd{s}^A_B)p(\sigma^B)\\
=&p(\sigma^A,\bd{s}^B_A)p(\sigma^B),
\end{align*}
where $\bd{s}^A_B= s^A_B(\sigma)$ and $\bd{s}^B_A=s^B_A(\sigma)$, for all $\sigma \in \sS^V_n$. The second equality holds because $\sigma^A_B(v)=\bd{s}^A_B(\sigma^A(v))$ for all $v \in A$, and thus $\event{\sigma^A_B}=\event{\sigma^A,\bd{s}^A_B}$. The third equality holds because the ranks in $\bd{s}^B_A$ are uniquely determined by the ranks in $\bd{s}^A_B$ and thus $\event{\bd{s}^A_B}=\event{\bd{s}^B_A}$.

Then, by Definition~\ref{def:factorized.push}, we have
\begin{align*}
h(\sigma)=& g(\sigma^A)g(\sigma^B_A)\\
=& g(\sigma^A)g(\sigma^B,\bd{s}^B_A)\\
=& p(\sigma^A)p(\sigma^B)p(\bd{s}^B_A),\\
\end{align*}
where $\bd{s}^B_A=s^B_A(\sigma)$, for all $\sigma \in \sS^V_n$. The second equality holds because $\sigma^B_A(b)=\bd{s}^B_A(\sigma^B(b))$ for all $b \in B$.

ii) $h= \argmin_{q \in \RSM} D_{KL}(p||q)$:

By (i) $h \in \RSM$ and $h(\sigma)=p(\sigma^A)p(\sigma^B)p(\bd{s}^B_A)$.

The Kullback-Leibler divergence between $p$ and $g$ is:
\[D_{KL}(p||g)= \sum_{\sigma \in \sS^V_n} p(\sigma) \ln \frac{p(\sigma)}{g(\sigma)}.\]
Minimizing the divergence is equivalent to maximizing the expected log-likelihood,
\[\argmin_{g \in \RSM}D_{KL}(p||g)= \argmax_{g \in \RSM} \sum_{\sigma \in \sS^V_n} p(\sigma) \ln g(\sigma).\]
Any $g \in \RSM$ factorizes as
\[g(\sigma)=g(\sigma^A)g(\sigma^B)g(s^B_A(\sigma)).\]

Using the factorization, we can decompose the objective function of the maximization problem additively:
\[\sum_{\sigma \in \sS^V_n} p(\sigma) \ln g(\sigma)= \sum_{\sigma \in \sS^V_n} p(\sigma) \ln g(\sigma^A) + \sum_{\sigma \in \sS^V_n} p(\sigma) \ln g(\sigma^B) + \sum_{\sigma \in \sS^V_n} p(\sigma) \ln g(s^B_A(\sigma)).\]

By grouping rankings in the events we have
\begin{align*}
\sum_{\sigma \in \sS^V_n} p(\sigma) \ln g(\sigma^A)=&\sum_{\sigma^A \in \sS^A_{|A|}}p(\sigma^A)\ln g(\sigma^A),\\
\sum_{\sigma \in \sS^V_n} p(\sigma) \ln g(\sigma^B)=&\sum_{\sigma^B \in \sS^B_{|B|}}p(\sigma^B)\ln g(\sigma^B),\\
\sum_{\sigma \in \sS^V_n} p(\sigma) \ln g(s^B_A(\sigma))=&\sum_{\bd{s}^B_A \in \binom{N}{|B|}}p(\bd{s}^B_A)\ln g(\bd{s}^B_A).
\end{align*}

Therefore, the RSM that minimizes the Kullback-Leibler divergence with respect to $p$ is $h$, where
\[
h(\sigma)=p(\sigma^A)p(\sigma^B)p(s^B_A(\sigma)),
\]
for all $\sigma \in \sS^V_n$.
\end{proof}


\end{document}